\documentclass[APA,STIX1COL]{ION-APA_Template}
\usepackage{tabularx}

\usepackage{amsmath,amsfonts}
\usepackage{balance}
\usepackage{algorithm}
\usepackage{algpseudocode}

\makeatletter
\def\NAT@aysep{,}
\makeatother

\articletype{Original Article}%

\received{--} 
\revised{--}
\accepted{--}

\begin{document}

\title{ICET Online Accuracy Characterization for Geometry-Based Laser Scan Matching} 

\author[1]{Matthew McDermott*}

\author[1]{Jason Rife}

\authormark{McDermott \textsc{et al}}

\address[1]{\orgdiv{Department of Mechanical Engineering}, \orgname{Tufts University}, \orgaddress{\state{Massachusetts}, \country{USA}}}



\corres{Matthew McDermott, Tufts University, 419 Boston Ave. Medford MA 02155 \email{matthew.mcdermott@tufts.edu}}


\abstract[Summary]{ 
Distribution-to-Distribution (D2D) point cloud registration algorithms are fast, interpretable, and perform well in unstructured environments.
Unfortunately, existing strategies for predicting solution error for these methods are overly optimistic, 
particularly in regions containing large or extended physical objects.
In this paper we introduce the \textit{Iterative Closest Ellipsoidal Transform} (ICET), a novel 3D LIDAR scan-matching algorithm that re-envisions NDT in order to provide robust accuracy prediction from first principles.
Like NDT, ICET subdivides a LIDAR scan into voxels in order to analyze complex scenes by considering many smaller local point distributions, however, ICET assesses the voxel distribution to distinguish random noise from deterministic structure. ICET then uses a weighted least-squares formulation to incorporate this noise/structure distinction into computing a localization solution and predicting the solution-error covariance. In order to demonstrate the reasonableness of our accuracy predictions, we verify 3D ICET 
in three LIDAR tests involving
real-world automotive data, high-fidelity 
simulated trajectories, and simulated corner-case scenes. For each test, 
ICET consistently performs scan matching with sub-centimeter accuracy. This level of accuracy, combined with the fact that the algorithm is fully interpretable, make it well suited for safety-critical transportation applications. Code is available at \url{https://github.com/mcdermatt/ICET}

    }

\keywords{Localization, Scan Registration, SLAM, LIDAR} 

\jnlcitation{\cname{%
\author{McDermott M.}, 
\author{Rife J.}}, 
\ctitle{ICET Online Accuracy Characterization for Geometry-Based Laser Scan Matching}. }

\maketitle

\footnotetext{\textbf{Abbreviations:} D2D, Distribution-to-Distribution; ICP, Iterative Closest Point; DNN, Deep Neural Network}

\section{Introduction}\label{Introduction}

Scan matching is the task of finding the rigid transformation that best aligns two point clouds. Scan matching can be used to gain insight on the motion of a vehicle and to build HD Maps of an environment. In this paper we introduce the Iterative Closest Ellipsoidal Transform (ICET), a novel algorithm that is capable of robust point-cloud registration and accurate prediction of its own solution error as a function of scene geometry. Accurately modeling solution accuracy is useful for sensor fusion, and necessary for safety-of-life navigation tasks. Additionally, ICET can identify degenerate cases where there is insufficient information to constrain one or more components of its solution and suppress the corresponding output without adversely impacting solution accuracy about other axis.

At its core, ICET uses a weighted least-squares approach for scan matching and accuracy prediction. Like the popular 3D-NDT D2D \citep{D2DNDT} algorithm, which obtains its solution through gradient-descent optimization, our approach also uses a voxel grid to subdivide a full LIDAR scan into smaller point clusters, which can be aligned between sequential scans to infer relative motion. 

An important difference is that our approach further analyzes the point distribution within each voxel to distinguish between noise 
and deterministic structure.
ICET addresses this noise/structure distinction and thereby excludes ambiguous measurements and
generates a better approximation of the measurement-noise covariance matrix. The measurement-noise covariance is then used in two ways: as a weighting matrix to enhance solution accuracy and as an input for predicting the solution-error covariance.

The key contribution of the algorithm is combining noise/structure differentiation with covariance prediction in voxel-based LIDAR scan matching. We believe this characteristic of ICET is new in the research literature, although elements have been presented in the past. For instance, though solution-error covariance is not generally predicted in most NDT implementations, an approach has been presented previously by \cite{D2DNDT}, who extended Censi's method \citep{censi} to the specific case of an NDT cost function. However, this approach was applied to voxel measurement distributions with no consideration for the differences between random measurement noise and the systematic spread of points associated with object shape.  Conversely, the NDT variant implemented by Einhorn et al. seeks increased accuracy by  
suppressing measurements except in the the voxel distribution's most compact 
principal-axis 
direction
\citep{EINHORN201528}; however, their approach does not tie ambiguity mitigation to covariance estimation. 
In a similar vein, DMLO combines geometric methods with an Artificial Neural Network (ANN) to account for distribution shape, but not in the context of predicting solution-error covariance \citep{DMLO}. Putting the noise/structure distinction in the context of covariance prediction is particlarly important for driving applications, where tunnels and highways with long flat walls can introduce ambiguities for voxel-based scan matching algorithms and cause bad localization performance~\citep{CMUtunnel, wen2018performance}. 

The cornerstone of the ICET implementation is its use of a standard least-squares methodology for deriving motion states from LIDAR data.  Least-squares implementations of NDT \citep{LS3DNDT} are significantly simpler than traditional Point-to-Distribution (P2D) implementations of NDT, which obtain solutions through gradient-descent optimization. Least-squares techniques \citep{Simon} are widely known, easy to implement, and computationally efficient. 
Importantly, the relationship between the input measurement-noise covariance and the output solution-error covariance is well understood for least-squares methods. This relationship relies on the assumption that the noise inputs to least-squares processing are strictly random in nature, as is assumed in traditional NDT \citep{Biber,D2DNDT}. In reality, the spread of points within each voxel is due to a combination of random noise and deterministic structure. The key point of this paper is that the later must be suppressed in order to achieve robust error characterization.

The interpretable nature of the ICET algorithm and  its enhanced noise modeling make it particularly well suited for safety-of-life applications like automated driving. Developing a rigorous safety case usually requires the use of transparent algorithms, to identify and bound the risks of rare but potentially hazardous situations, which represent a risk to navigation integrity~\citep{zhu2022integrity,bhamidipati2022robust}. By comparison, methods that rely on machine learning, like LO-Net \citep{Li} and VertexNet \citep{vertexnet}, are not currently interpretable, which makes them difficult to characterize through a rigorous safety case. With regard to safety certification, voxel-based methods like NDT and ICET are also advantageous in that they do not introduce significant risks associated with data association (or \textit{correspondence}) between features drawn from different LIDAR scans. As pointed out by \cite{hassani2018lidar}, data association issues make the safety case very challenging when matching landmarks, features, or points from LIDAR scans, as in the broad class of Iterative Closest Point (ICP) algorithms such as Generalized-ICP \citep{generalizedICP}, LOAM \citep{LOAM}, and I-LOAM \citep{Park}.

The remainder of this paper introduces a 3D formulation of ICET and verifies it through simulation. The next section, Section 2, provides additional background related to noise-modeling for voxel-level LIDAR point distributions. Section 3 provides the key equations for the 3D formulation of ICET. Section 4 evaluates the performance of ICET against NDT on a real LIDAR dataset. This analysis provides a proof-of-concept that ICET processes real-world data effectively; however, the GPS reference data (cm-level errors between epochs) is significantly less accurate than the LIDAR (mm-level errors between scans). As such, section 5 performs a similar test using simulated LIDAR, where perfect ground truth is available. 
Section 6 contains a final corner-case simulation to show how ICET identifies rare instances where there is insufficient information to constrain solutions about all axes (as in the case of traveling through a straight tunnel); in these cases, ICET provides as complete a solution as possible given detected ambiguity. Finally, section 7 discusses limitations of ICET and possible future enhancements.


\section{Defining the Problem}\label{Problem}

\begin{figure}[b]
\centering
\includegraphics[width=6.0in]{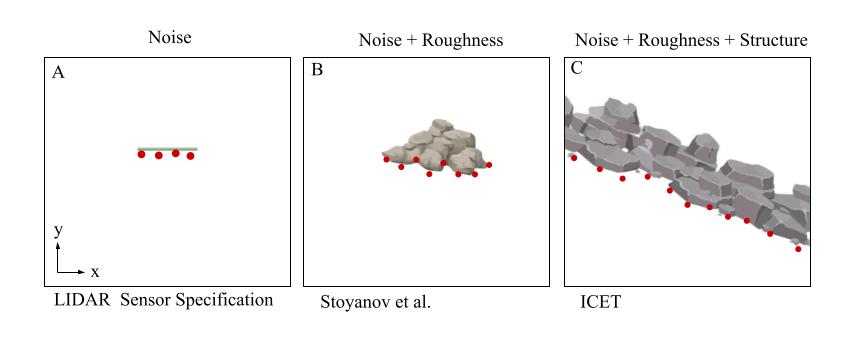}
\caption{Voxel point distributions featuring (A) sensor noise, (B) surface roughness, and (C) extended surfaces }
\label{fig:noisevsstructure}
\end{figure}

This section visualizes the fundamental problem that motivates the ICET algorithm: separating deterministic object shape from random measurement noise, effects that together determine the distribution of LIDAR points across a voxel. 

In modeling voxel-level noise, one important consideration is intrinsic measurement noise, caused by the sensor itself. For example, errors are introduced by range-measurement circuitry and (for rotating units) by encoders that measure LIDAR orientation. Noise levels (e.g., one-sigma error) associated with intrinsic noise can be characterized in device specifications published by the manufacturer. 
As a visualization of intrinsic noise, consider the LIDAR points (red dots) in Figure \ref{fig:noisevsstructure}A. The surface is completely smooth, but LIDAR samples are scattered about the surface due to small errors in measuring time of flight. 

In modeling voxel-level noise, a second important consideration is discrete sampling on rough surfaces.  Local surface texture on the scale of the spatial resolution of the LIDAR results in measurement variations amplified by multipath
\footnote{On non-flat surfaces, the width of the LIDAR beam can result in multiple returns, or \textit{multipath} \citep{lidarMultipath}. As an example the Velodyne VLP-16 unit has a beam width of 8 cm at a range of 25 m~\citep{velodyne}. This Velodyne unit can be configured to capture either the first return or the strongest return. When viewing a corner or rough surface, toggling this setting adjusts the expected value of the range measurement.}.
Surface roughness results in additional apparent noise because the sampled locations  change continuously due to LIDAR motion. Because these effects are highly dependent on the nature of the surface creating the LIDAR reflections, these effects are difficult to characterize in a general way, so they are not characterized in the system specification. In Figure \ref{fig:noisevsstructure}B, a rock formation is observed with significant surface roughness, which amplifies the effective noise above the levels observed in Figure \ref{fig:noisevsstructure}A. 

In addition to these random effects, the physical shape of the reflecting surface also influences the LIDAR point distribution. When viewing small objects, as in Figure \ref{fig:noisevsstructure}A and \ref{fig:noisevsstructure}B, the distinction between random noise and deterministic object structure is not significant; however, when the object becomes wide, as in \ref{fig:noisevsstructure}C, then the object shape becomes significant, and the distinction between random noise and deterministic structure cannot be ignored.



The combination of random and deterministic factors makes it somewhat difficult to model  measurement noise.  The most straightforward approach is to use manufacturer specifications, but these  underestimate the actual measurement covariance \cite{glennie2010static}, presumably because they reflect only intrinsic noise (Figure \ref{fig:noisevsstructure}A) and not discrete sampling effects on rough surfaces (Figure \ref{fig:noisevsstructure}B). As an alternative, the noise covariance can be estimated statistically from the distribution of points in a voxel, as was done in \cite{D2DNDT}. Then the resulting covariance captures both intrinsic noise and discrete sampling effects (as shown in Figure \ref{fig:noisevsstructure}B); however, this statistical estimate also treats large-scale shape effects as noise (as visible in Figure \ref{fig:noisevsstructure}C). An artificially large estimate of the noise covariance results. (The computed covariance along the extended surface of Figure \ref{fig:noisevsstructure}C is essentially equal to the moment of inertia for a long rod.)

A compromise solution is to model the noise covariance based on the distribution of points in the voxel, but only after excluding distribution directions that stretch across the length of the voxel.  As long as the voxel is large enough to contain typical noise distributions, but not significantly larger, then the LIDAR point distribution will be dominated by random effects as long as extended directions are excluded. We take this approach in ICET.


\section{Implementation}\label{Implementation}
This section details the ICET algorithm, first covering the basic algorithm (Section 3.1), then discussing the noise-modeling process (Section 3.2), and closing with additional refinements needed to process real-world data (Section 3.3). The development is a refined and more complete version of our conference paper \citep{ICET}. Importantly, 
Section 3.2 details the paper's key contribution, which is related to suppressing extended-axes of the point-cloud distribution in order to provide a more reliable estimate of measurement noise. 


\subsection{ICET Least-Squares Solution in 3D}
ICET generates a solution vector that represents the translation and rotation of a new scan to best align with an earlier scan, assuming both scans capture roughly the same stationary (or near stationary) objects and terrain features. The  methodologies presented in this section are, for the most part, representative of those used in a wide range of existing NDT implementations.

In ICET, the solution vector \textbf{x} includes Cartesian translations $\{x,y,z\}$ and body-fixed XYZ Euler angles $\{\phi,\theta,\psi\}$ needed to align the scans. 

\begin{equation}\label{eq:xvec}
    \textbf{x} = \begin{bmatrix}
        x &
        y &
        z &
        \phi & 
        \theta & 
        \psi 
    \end{bmatrix}^T
\end{equation}

\noindent Combining the rotation and translation, it is possible to map the position vector $^{(i)}\mathbf{p}$ for each point $i$ in the new scan into the coordinate system of the earlier (or \textit{reference}) scan. The transformed point positions, labeled $^{(i)}\mathbf{q}$, are as follows.

\begin{equation}\label{eq:twoDtransform}
    ^{(i)}\textbf{q} = \textbf{R}
    {}^{(i)}\textbf{p} - 
    \begin{bmatrix}
        x &
        y &
        z 
    \end{bmatrix}^T
\end{equation}

\noindent In this equation, the matrix $\mathbf{R}$ is the 3D rotation matrix constructed from the roll, pitch, and yaw angles $\{\phi,\theta,\psi\}$.

\begin{equation}\label{eq:R3D}
    \mathbf{R} = 
    \begin{bmatrix}
        c_\theta c_\psi & s_\psi c_\phi + s_\phi s_\theta c_\psi & s_\phi s_\psi - s_\theta c_\phi c_\psi\\
        -s_\psi c_\theta & c_\phi c_\psi - s_\phi s_\theta s_\psi & s_\phi c_\psi + s_\theta s_\psi c_\phi \\
        s_\theta & -s_\phi c_\theta & c_\phi c_\theta \\
    \end{bmatrix} 
\end{equation}
\smallskip

Points are clustered using an arbitrarily shaped grid of $J$ voxels, centered on the location of the lidar. The grid is used because it mitigates data association fragility between scans, a limitation of early scan-matching algorithms such as the Iterative Closest Point algorithm \citep{Besl, Chen}. By assigning points to voxels, it is possible to compute a mean position ${}^{(j)}\mathbf{y} \in \mathbb{R}^{3\times1}$ for the new scan points and a mean position ${}^{(j)}\mathbf{y}_0 \in \mathbb{R}^{3\times1}$ for the reference scan points in each voxel~$j$. Similarly, covariance matrices ${}^{(j)}\mathbf{Q}$ and ${}^{(j)}\mathbf{Q}_0  \in \mathbb{R}^{3\times3}$ can be constructed for each voxel. Note that the $0$-subscript is used to identify reference-scan variables and distinguish them from new-scan variables. Once means ${}^{(j)}\mathbf{y}_0$ for each voxel in the reference scan are computed, they are concatenated into one large observation vector $\mathbf{y}_0 \in \mathbb{R}^{3J \times 1}$; the states $\mathbf{x}$ are then tuned so that the observation vector $\mathbf{y}_0$ is approximated as closely as possible by a similar concatenated vector constructed from the new scan $\mathbf{y}=\mathbf{h}(\mathbf{x}) \in \mathbb{R}^{3J \times 1}$, where the $\mathbf{h}(\mathbf{x})$ function is the vertical concatenation of $J$ repetitions of ${}^{(j)}\mathbf{h}(\mathbf{x}) \in \mathbb{R}^{3 \times 1}$, which are the new-scan voxel means computed after the new scan is transformed by~(\ref{eq:twoDtransform}).

ICET uses an iterative least-squares approach to estimate the state as $\hat{\mathbf{x}}$, leveraging the Taylor series expansion:

\begin{equation}\label{eq:taylor}
    \textbf{y}_0 = \textbf{h}(\hat{\textbf{x}}) + \textbf{H}\delta \textbf{x} + \mathcal{O}(\delta \textbf{x}^2)
\end{equation}

\noindent Starting with an initial guess of $\hat{\mathbf{x}}$, the above equation is solved assuming second and higher-order terms are small, to obtain the state correction $\delta \textbf{x}$. This Newton-Raphson operation is repeated to a reasonable convergence criterion, to obtain a final least-squares state estimate $\hat{\mathbf{x}}$. Though the Newton-Raphson method does not guarantee global convergence, this solution approach and related gradient-descent algorithms are widely used in LIDAR scan-matching, and countless implementations demonstrate their reliable convergence in practice.


Solving (\ref{eq:taylor}) requires that the Jacobian be evaluated for each iteration. In 3D, the Jacobian matrix $\mathbf{H} \in \mathbb{R}^{3J \times 6}$ can be constructed as a vertical concatenation of submatrices ${}^{(j)}\mathbf{H} \in \mathbb{R}^{3 \times 6}$, defined as follows for each voxel $j$.


\begin{equation}\label{eq:H}
    {}^{(j)}\textbf{H} = 
    \Bigg{[}
    \begin{matrix}
        -1 & 0 & 0 \\
        0 & -1 & 0 \\
        0 & 0 & -1 \\
    \end{matrix}
    \Bigg{|}
    \begin{matrix} 
        {}^{(j)}\mathbf{H}_\phi & {}^{(j)}\mathbf{H}_\theta & {}^{(j)}\mathbf{H}_\psi 
    \end{matrix}
    \Bigg{]}
\end{equation}

\noindent Here the vectors ${}^{(j)}\mathbf{H}_i$ are defined ${}^{(j)}\mathbf{H}_i = \frac{\delta \textbf{R}}{\delta i}\, {}^{(j)}\textbf{y}_p$, where ${}^{(j)}\mathbf{y}_p$ are the new-scan voxel-mean vectors in their original coordinates system (i.e., the mean of the ${}^{(i)}\mathbf{p}$ for all points $i$ inside the voxel $j$), and where the rotation matrix derivatives are the following matrices in $\mathbb{R}^{3 \times 3}$.
Similar rotation-matrix derivatives are used widely in scan matching, for example by \cite{hong2017probabilistic}.

\begin{equation}\label{eq:dRdphi}
    \frac{\delta \textbf{R}}{\delta \phi} = 
    \begin{bmatrix}
        0 & -s_\psi s_\phi + c_\phi s_\theta c_\psi & c_\phi s_\psi + s_\theta s_\phi c_\psi \\
        0 & -s_\phi c_\psi - c_\phi s_\theta s_\psi & c_\phi c_\phi - s_\theta s_\psi s_\phi  \\
        0 & -c_\phi c_\theta & -s_\phi c_\theta  \\
    \end{bmatrix} 
\end{equation}

\begin{equation}\label{eq:dRdtheta}
    \frac{\delta \textbf{R}}{\delta \theta} = 
    \begin{bmatrix}
        -s_\theta c_\psi & c_\theta s_\phi c_\psi & -c_\theta c_\phi c_\psi  \\
        s_\psi s_\theta  & -c_\theta s_\phi s_\psi & c_\theta s_\psi c_\phi \\
        c_\theta & s_\phi s_\theta & -s_\theta c_\phi  \\
    \end{bmatrix}
\end{equation}

\begin{equation}\label{eq:dRdpsi}
    \frac{\delta \textbf{R}}{\delta \psi} = 
    \begin{bmatrix}
        -c_\theta s_\psi & c_\psi c_\phi - s_\phi s_\theta s_\psi & c_\psi s_\phi + s_\theta c_\phi s_\psi  \\
        -c_\psi c_\theta & -s_\psi c_\phi - s_\phi s_\theta c_\psi & s_\theta c_\psi c_\phi-s_\phi s_\psi   \\
        0  & 0 & 0 \\
    \end{bmatrix}
\end{equation}
\smallskip

In solving (\ref{eq:taylor}) we use a weighted-least squares (WLS) formulation with the weighting matrix $\mathbf{W}$ constructed from the voxel-mean covariance matrices.  That is $\mathbf{W}=\mathbf{\Sigma}^{-1}$ where $\mathbf{\Sigma}$ is block diagonal with each block covariance ${}^{(j)}\mathbf{\Sigma}$ corresponding to one voxel~$j$. The block covariances reflect the uncertainty in the difference between the voxel means.  


\begin{equation}\label{eq:sigmaDef}
    {}^{(j)}\mathbf{\Sigma} =
    \mathrm{E}\Big[\big({}^{(j)}\mathbf{h}(\hat{\mathbf{x}})-{}^{(j)}\mathbf{y}_0 \big) \big({}^{(j)}\mathbf{h}(\hat{\mathbf{x}})-{}^{(j)}\mathbf{y}_0  \big)^T \Big] \end{equation}

\noindent Once $\mathbf{\Sigma}$ is estimated and inverted to obtain $\mathbf{W}$, the standard WLS solution can be computed iteratively with the following equation~\citep{Simon}.

\begin{equation}\label{eq:WLSsoln}
    \delta\mathbf{x} = (\mathbf{H}^T\mathbf{W}\mathbf{H})^{-1}\mathbf{H}^T\mathbf{W}\big(\mathbf{y}_0-\mathbf{h}(\hat{\mathbf{x}})\big)
\end{equation}

\noindent After each iteration of (\ref{eq:WLSsoln}), the state estimate is updated ($\hat{\textbf{x}} \rightarrow \hat{\textbf{x}}+\delta\textbf{x}$) until a reasonable convergence criterion is satisfied.

The WLS formulation employed by ICET makes accuracy prediction straightforward in a computational sense. Applying the standard WLS formulation \citep{Simon},
the state-error covariance $\mathbf{P} \in \mathbb{R}^{6 \times 6}$ is

\begin{equation}\label{eq:Pmatrix}
    \mathbf{P} = (\mathbf{H}^T\mathbf{WH})^{-1}.
\end{equation}

An important detail in our implementation of ICET is that we use computationally efficient forms to obtain the solution (\ref{eq:WLSsoln}) and the covariance (\ref{eq:Pmatrix}). Given the block diagonal structure of the variables, it is possible to frame (\ref{eq:WLSsoln}) in an equivalent but more computationally efficient manner, by eliminating many multiplications by zero.  The equivalent form recognizes that

\begin{equation}\label{eq:WLSblock}
    (\mathbf{H}^T\mathbf{W}\mathbf{H})^{-1}=\sum_{j\in 1:J}{\Big[{}^{(j)}\mathbf{H}^T
    \;{}^{(j)}\mathbf{\Sigma}^{-1}
    \;{}^{(j)}\mathbf{H}
    \Big]}
\end{equation}

\noindent and 

\begin{equation}\label{eq:WLSblock2}
    \mathbf{H}^T\mathbf{W}\big(\mathbf{y}_0-\mathbf{h}(\hat{\mathbf{x}})\big)
    =\sum_{j\in 1:J}{\Big[{}^{(j)}\mathbf{H}^T
    \;{}^{(j)}\mathbf{\Sigma}^{-1}
    \;\big({}^{(j)}\mathbf{y}_0-{}^{(j)}\mathbf{h}(\hat{\mathbf{x}})\big)
    \Big]}.
\end{equation}

\noindent In addition to speeding computation, the evaluation of (\ref{eq:WLSsoln}) using (\ref{eq:WLSblock}) and (\ref{eq:WLSblock2}) is also less memory intensive, because sequentially processing small $\mathbb{R}^{6\times6}$ blocks avoids the construction of the arbitrarily large matrices $\mathbf{H}$ and $\mathbf{W}$. In the process, the value of the covariance matrix (\ref{eq:Pmatrix}) is automatically obtained from (\ref{eq:WLSblock}). Accordingly, the ICET covariance-estimate is significantly easier to compute than the NDT covariance-estimate derived in \cite{D2DNDT}.

\subsection{Refinement of Measurement Covariance \label{app1}}

Though computationally straightforward to evaluate, the solution covariance (\ref{eq:Pmatrix}) 
is only as reliable as the estimated measurement-covariances $\mathbf{\Sigma}$.  In order to obtain a good estimate of $\mathbf{\Sigma}$, it is important to think carefully about the random and deterministic factors that distribute LIDAR samples within a voxel, as  discussed in Section 2.  In this section we present our main contribution, the algorithmic step that identifies and excludes ambiguous measurements from extended objects, such as the example object shown in Figure \ref{fig:noisevsstructure}B.

In ICET, our approach for modeling measurement noise within each voxel is to start with the sample covariance, and then to exclude directions where the point distribution is dominated by deterministic structure.  This approach allows us to capture random effects (measurement noise and discrete sampling) without \textit{a prior} modeling. Residual deterministic effects remain (e.g., the structural distributions for small objects), and these add some amount of additional conservatism; however, ambiguous information associated with large objects is removed, which greatly improves statistical estimation of $\mathbf{\Sigma}$.

Our implementation, like that of \cite{D2DNDT}, starts by computing the sample covariance of point locations within each voxel.  We label the sample covariance ${}^{(j)}\mathbf{Q}$ in the new scan and ${}^{(j)}\mathbf{Q}_0$ in the reference scan. Our approach begins to diverge from \citep{D2DNDT} in that ICET uses the mean as an observable for each voxel; as such, the point-location covariance must be converted to the covariance of the mean. For a random set of independent samples, the central limit theorem~\citep{nist2006sematech} implies that the covariance of the sample mean vector is the true covariance divided by the number of samples, which is ${}^{(j)}N$ for the new scan and ${}^{(j)}N_0$ for the reference scan. Since the actual observable is the mean \textit{difference} between the new and reference scans, as indicated by (\ref{eq:sigmaDef}) and (\ref{eq:WLSsoln}), the contributions of both scans must be considered. By extension, assuming independent samples within each scan and between scans, the measurement covariance ${}^{(j)}\mathbf{\Sigma}$ for each voxel $j$ is 

\begin{equation}\label{eq:sigma}
    {}^{(j)}\mathbf{\Sigma} \approx
    \frac{{}^{(j)}\mathbf{Q}}{{}^{(j)}N} + 
    \frac{{}^{(j)}\mathbf{Q}_0}{{}^{(j)}N_0}.
\end{equation}

\noindent A minor subtlety is that the above formula substitutes the sample covariance for the true covariance (which is not known); the distinction is not significant given that the number of samples in each voxel is sufficiently high. To this end we introduce a minimum-sample count (50 points per voxel in this paper) below which the voxel is excluded from computation.

Our next step is to exclude directions from the computation where the sample variance is unrealistically high. This step further distinguishes ICET from prior work that has attempted to characterize state-error in voxel-based scan matching. Specifically, ICET removes extended distribution directions using an eigentransformation of the point-location covariance matrix for each voxel $j$.  The eigentransformation generates a unitary eigenvector matrix ${}^{(j)}\mathbf{U} \in \mathbb{R}^{3 \times 3}$ and a diagonal eigenvalue matrix ${}^{(j)}\mathbf{\Lambda} \in \mathbb{R}^{3 \times 3}$ for each voxel. 

\begin{equation}\label{eq:eigen1}
    {}^{(j)}\mathbf{Q}_0 = {}^{(j)}\mathbf{U} \,\,\, {}^{(j)}\mathbf{\Lambda}\,\,\, {}^{(j)}\mathbf{U}^T
\end{equation}

\noindent The eigenvectors describe the principal-axis directions, whereas the eigenvalues describe the principal-axis lengths (squared) for the covariance ellipsoid.  Directions where the distribution of points is dominated by deterministic structure are identified as directions where the eigenvalue is large.

Note that the eigentransformation is applied to ${}^{(j)}\mathbf{Q}_0$, which describes the reference scan. Deterministic structure is embedded not only in 
${}^{(j)}\mathbf{Q}_0$, of course, but also in ${}^{(j)}\mathbf{Q}$ and ${}^{(j)}\mathbf{\Sigma}$. The reason to focus on ${}^{(j)}\mathbf{Q}_0$ is computational efficiency, noting that the voxel boundaries are defined for the reference scan, so the LIDAR points are consistently assigned to the same voxel in the reference scan throughout the WLS iterations of (\ref{eq:WLSsoln}) to convergence. By comparison, each iteration re-applies~(\ref{eq:twoDtransform}) to transform the new scan differently on to the voxel grid, so point associations for the new scan change after each WLS iteration. By focusing on the reference scan, the eigentransformation is computed infrequently, only when the reference image is updated.

\begin{figure}[h]
\centering
\includegraphics[width=3.0in]{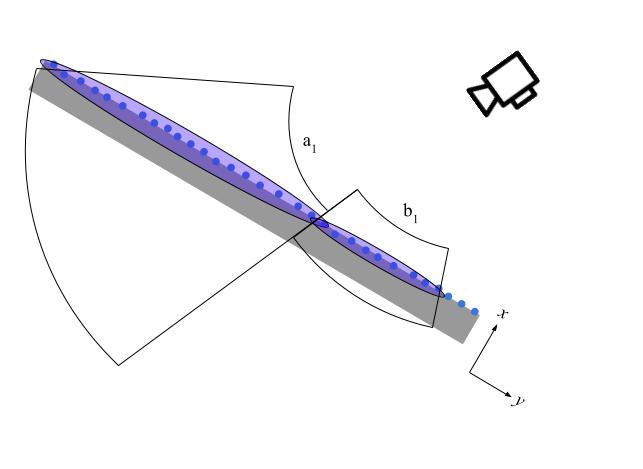} 
\caption{
Illustration of a wall (seen from above) passing through two rounded voxels.  LIDAR measurements of the surface are shown as dots.  The mean and covariance matrix for the scan points in each voxel are indicated by the shaded ellipsoids.} 
\label{fig:SP}
\end{figure}

The next step is to define the eigenvalue cutoff beyond which corresponding eigenvector directions are excluded from the solution. In this regard, our basic premise is that some inflation of the sample covariance due to deterministic structure is acceptable, but that a significant problem can result if the point distribution appears to extend across the voxel, exiting the voxel through two of its boundaries. Voxels are typically configured to be wide enough (20 cm or wider across) to contain essentially all random noise; as such voxel width provides a coarse but effective reasonableness check to detect large distribution widths caused by physical structure.  This concept is illustrated by Figure~\ref{fig:SP}, which depicts a birds-eye view of a wall (gray rectangle) and the distribution of LIDAR returns(blue dots) generated by the wall. The figure depicts two rounded voxels, which are characteristic of the voxel shapes used in ICET (see next section). A covariance estimate can be constructed from the LIDAR points in each voxel. Notably the covariance ellipse extends entirely across each voxel, from one voxel face to another.

To a human observer, it is evident from the figure that the distribution of LIDAR points along the wall is dominated by deterministic object structure and not random effects; however, the statistical estimation process of (\ref{eq:sigma}) is blind to the distinction and infers incorrectly that the distribution of points along the wall surface is random. This inference is inherently problematic because the voxel-mean remains centered along the segment passing through the voxel even for small perturbations of the LIDAR unit's location, which means that the mean conveys no useful information for localization in the along-wall direction. Using an along-wall measurement in the least-squares solution (\ref{eq:WLSsoln}) injects error but no signal. In this sense, both voxels shown in the figure are problematic unless the along-wall measurement is suppressed; the effect is even more severe for the righthand voxel than the left, because the distribution for the righthand voxel appears narrower in the along-wall direction, implying a lower variance and, accordingly, a higher weighting in the solution.


This ambiguity issue is a specific example of the aperture problem \citep{shimojo1989occlusion}, which is well known in the computer vision literature, but which has not been discussed in the LIDAR literature prior to the work of \cite{ICET}.

Introducing a cutoff that limits the magnitude of the eigenvalues for ${}^{(j)}\mathbf{Q}_0$ both caps unrealistic distributions and addresses the ambiguity introduced when relatively flat objects stretch across an entire voxel.  We do not recommend a fixed threshold cutoff, because the amount of the large object contained inside a voxel depends on its orientation within the voxel, as shown in Figure~\ref{fig:SP}. Rather, in this paper, we recommend that the cutoff be framed in terms of the relationship between the point distribution and the voxel boundaries. Namely, we propose that if the point distribution along any principal-axis direction extends past two voxel boundaries, then the associated principal-axis direction should be excluded from the solution.

\begin{figure}[h]
\centering
\includegraphics[width=4.0in]{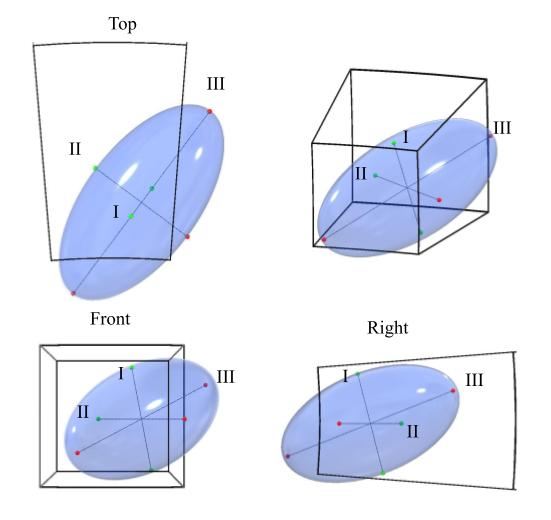}
\caption{Sigma-point refinement for identification of extended surfaces within a single voxel. A covariance ellipse is shown in relation to a wireframe voxel. The sigma points along the ellipsoid's three principal axes (labeled I, II, and III), are identified with small dots. Only axis III is eliminated by our sigma-point exclusion text, since both of its sigma points lie outside the voxel boundaries.}
\label{fig:UKF}
\end{figure}

Our implementation of this cutoff is inspired by sigma-point filters, also known as Unscented Kalman Filters or UKFs~\citep{julier2004unscented,nonlinearKalman}. 
They model a covariance matrix by placing two points along each principal axis (or eigenvector direction), spaced symmetrically around the mean.  In our case, we represent the covariance matrix with six test points (two for each direction in 3D space), placed along each eigenvector axis at a distance of 2$\sigma$ on either side of the mean, where $\sigma$ is used here to refer to the square-root of the eigenvalue associated with the eigenvector. If both sigma points for any given eigenvector lie outside the voxel, then ICET suppresses measurement information in that direction. Axes with only a single test point outside the voxel boundary are included in the solution, as they still contribute to localization in a meaningful way.  The sigma-point test is illustrated in Figure~\ref{fig:UKF}.

Figure \ref{fig:UKF} illustrates the pruning process. The figure shows only the sample covariance (blue ellipsoid) and the voxel boundaries (wireframe). The LIDAR points used to generate the covariance are not shown. The three principal axes of the ellipsoid are labeled I, II, and III, and the sigma points are indicated as dots along the length of each axis. The figure uses three orthographic views (top, front, right) and one isometric projection to show the sigma points from different angles. For principal axis I, both sigma points lie inside the voxel; for principal axis II, one sigma point lies inside and one outside; for principal axis III, both sigma points lie outside. As such, the pruning algorithm will only suppress measurements in the direction of principal axis III.

Pruning is applied before computing (\ref{eq:WLSsoln}), so that eigenvector axes excluded by the test can be removed from the WLS solution.
The specific process for removing extended directions is, for each voxel $j$, to introduce a projection matrix ${}^{(j)}\mathbf{L}$, which preserves only the eigenvector directions associated with compact distribution axes (which do not extend outside the confines of their respective cell boundaries). The projection removes those directions from (\ref{eq:taylor}), one voxel at a time. The modified residual vector ${}^{(j)} \Delta \tilde{\mathbf{y}}$ for each voxel $j$ is:

\begin{equation}\label{eq:dz}
    {}^{(j)} \Delta \tilde{\mathbf{y}} = {}^{(j)}\textbf{L}\: {}^{(j)}\textbf{U}^T ({}^{(j)}\textbf{y}_0 - {}^{(j)}\textbf{y})
\end{equation}

\noindent Concatenating the modified residual vectors together, the WLS solution can be computed excluding data along extended surfaces. 
Unlike the strategy employed by \citep{EINHORN201528} that preserves residuals along only the most compact distribution axis, our approach has the flexibility to capture any sufficiently compact component of a solution vector. For example, a tall and narrow world feature, such as a lamppost provides information in two directions and is ambiguous in one (along the post's vertical axis).  Our sigma-point refinement would preserve two axes in the case of the lamppost. By contrast, for a flat wall, the sigma-point refinement would preserve only one direction (normal to the wall) and reject the other two. Most importantly, though, our innovation is to couple axis-exclusion with covariance estimation, to enhance ICET's ability to estimate the solution covariance.

\subsection{Additional Solution Conditioning Steps \label{app2}}

Three further considerations are relevant to enhancing ICET results: defining voxel shape, checking solution rank, and detecting moving-objects.

\begin{figure}[b]
\centering
\includegraphics[width=7.0in]{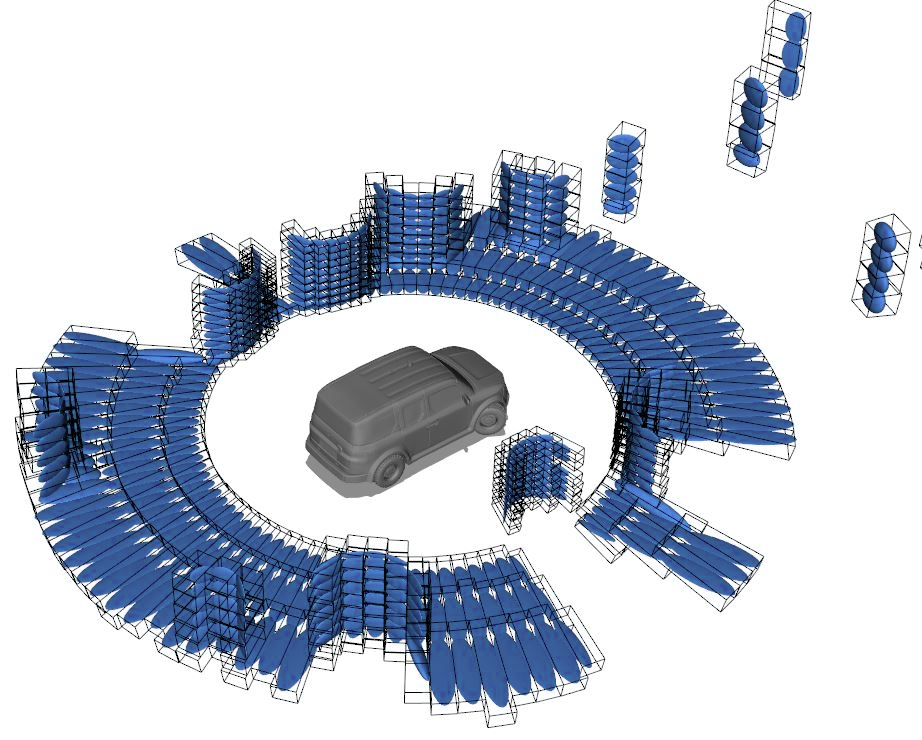} 
\caption{Spherical voxels with adaptive radial binning. Point distributions within each cell are shown as ellipsoids.}
\label{fig:shadow}
\end{figure}

With regard to voxel definition, the current ICET implementation uses a spherical voxel grid, with one-voxel radial depth in each azimuth and elevation direction.  The grid is generated by a shadow-mitigation algorithm that we have introduced as a pre-processing step for NDT and ICET \citep{shadowing}.  An illustration of the grid is shown in Figure \ref{fig:shadow}.  The value of spherical grids for LIDAR processing has also been recognized in other recent publications (\cite{hassani2021new}). 
For our application, spherical coordinates are particularly advantageous in that the voxels are aligned in the same direction in which discrete sampling occurs (i.e., radially away from the sensor). This allows voxel resolution to be increased in the azimuthal and elevation directions (to decrease the impact of shape effects) while retaining the ability to stretch in the radial direction as needed to properly characterize porous substrates (like loose foliage). 

With regard to solution rank, when some measurements directions are excluded, it is important to ensure that the matrix inverse in (\ref{eq:WLSsoln}) can be computed. Implementing a condition-number check can ensure that the solution is full rank (in a computational sense) and that the inverse can be computed. We implement a condition-number test by performing an eigendecomposion on the inverse of (\ref{eq:Pmatrix}) to obtain a unitary eigenvector matrix $\mathbf{U}_2\in \mathbb{R}^{6 \times 6}$ and eigenvalue matrix, $\mathbf{\Lambda}_2 \in \mathbb{R}^{6 \times 6}$

\begin{equation}\label{eq:cond1}
    \mathbf{H}^T\mathbf{WH} = \mathbf{U}_2 \mathbf{\Lambda}_2 \mathbf{U}^T_2
\end{equation}

\noindent
The condition number can be calculated as the ratio of the largest to the smallest eigenvalue in $\mathbf{\Lambda}_2$. In our testing, we found a suitable threshold $T_{cond}$ to be $5\times10^4$. If the condition number is below the threshold, a reliable solution can be generated.

If the condition number is larger than the threshold, a solution can still be computed in a reduced-dimension solution space. The dimension reduction process is to start with the smallest eigenvalue $\lambda_0$ and remove the associated eigenvector if the condition-number exceeds $T_{cond}$. The test is then repeated to remove eigenvector directions from the solution until the condition-number test is passed. This process allows for a valid solution in extreme cases when the same measurement direction is suppressed by the sigma-point test across in nearly all voxels, as might happen in the along-track direction for a straight, featureless tunnel.



With regard to moving-object detection, we use classical outlier rejection to detect and exclude voxels with large discrepancies between sequential frames, as caused by moving cars and pedestrians. This is achieved by first allowing the algorithm to converge on a preliminary solution. Once an initial registration is achieved, voxels are flagged in which the residual distance ${}^{(j)}\mathbf{y}_0-{}^{(j)}\mathbf{h}\big(\hat{\mathbf{x}}\big)$ exceeds a threshold $T_{mod}$. Flagged voxels are excluded, and the algorithm is run again to produce a final estimate. We set the threshold to be $T_{mod} = 5$ cm, which is high (about $5\sigma$) compared to typical measurement errors.

The full algorithm is summarized as pseudocode in the table labeled Algorithm \ref{tab:algo}.

\begin{algorithm} [t]
    \begin{algorithmic}[1]
    \caption{ICET Scan Registration}
    \label{tab:algo}
    \State Initialize new point cloud $K$ and reference point cloud $K_0$
    \State Define voxels $j \in (1,J)$ using a spherical grid with one voxel in each look direction \citep{shadowing}
    \For{each voxel $j$}
        \State Calculate mean $^{(j)}\mathbf{\mu}_0$ and covariance $^{(j)}\mathbf{Q}_0$ of points from reference scan $K_0$
        \State Perform eigendecomposition of  $^{(j)}Q_0$
        \State Apply sigma-point test, excluding any eigenvector direction with two test points outside voxel. 
    \EndFor
    \For{iteration $i$ until convergence}
        \State Transform new scan points $k \in K$ using (\ref{eq:twoDtransform})
        \State Associate transformed points with a voxel $j$
        \For{each voxel $j$}
            \State Calculate $^{(j)}\mathbf{\mu}$ and $^{(j)}\mathbf{Q}$ for new scan $K$
            \State Build measurement vector, applying eigenvector-direction exclusions from sigma-point test with (\ref{eq:dz})
            \State Conduct eigen-decomposition with (\ref{eq:cond1}) and eliminate solution directions to satisfy condition-number requirement 
            \State Compute state correction $\delta\textbf{x}$ using (\ref{eq:WLSsoln})
            \State Update states $\hat{\textbf{x}} \rightarrow \hat{\textbf{x}}+\delta\textbf{x}$
        \EndFor
\EndFor
\State Perform condition number check
\State Perform moving-object detection by flagging voxels with ${}^{(j)}\mathbf{y}_0-{}^{(j)}\mathbf{h}\big(\hat{\mathbf{x}}\big) > T_{mod}$
\State Exclude flagged voxels and repeat lines 9-17 
\end{algorithmic}
\end{algorithm}

\newpage
\section{Verification Testing: Real LIDAR Dataset}\label{reallidar}

\begin{figure}[h]
\centering
\includegraphics[width=2.0in]{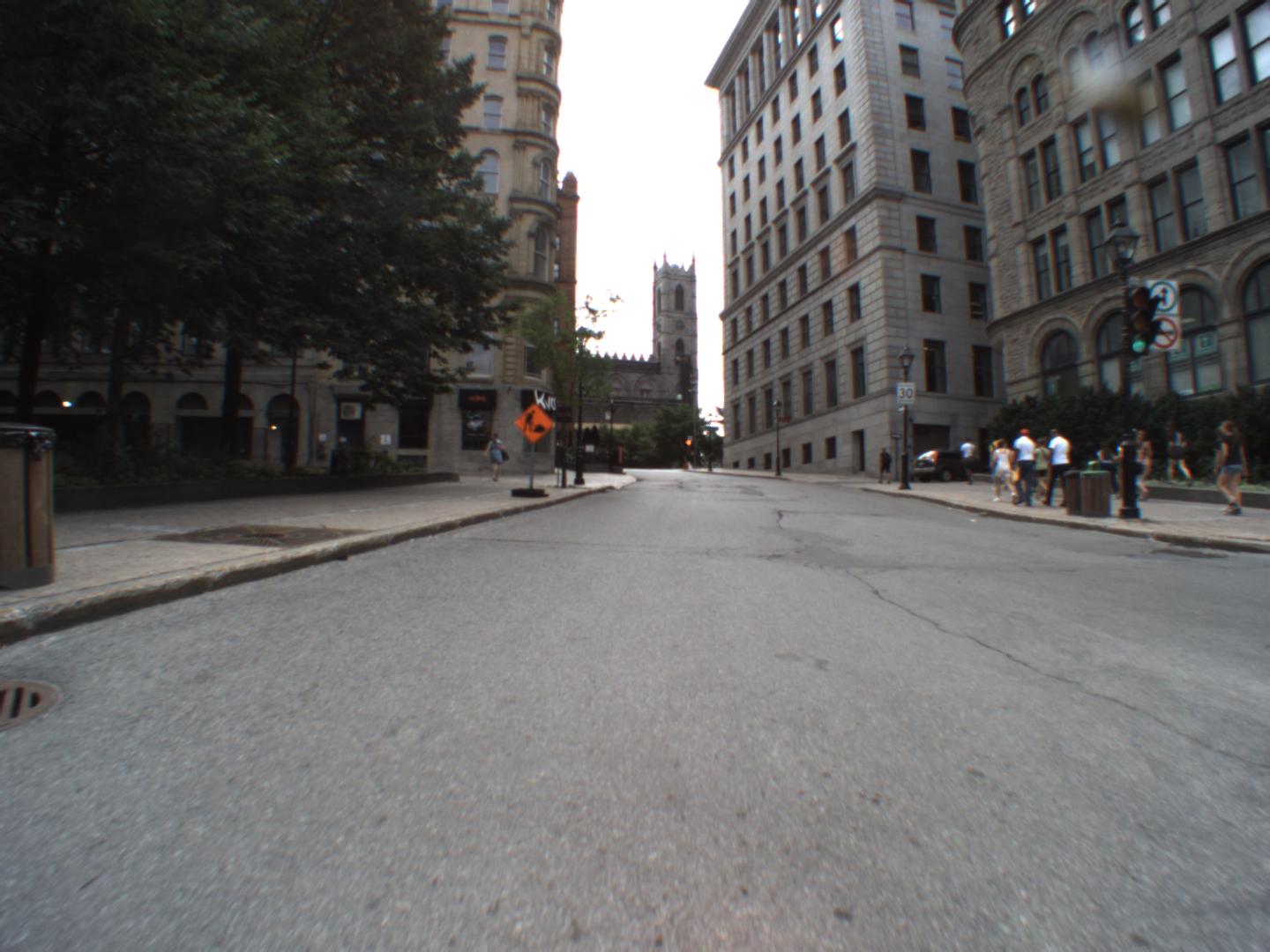}
\includegraphics[width=2.0in]{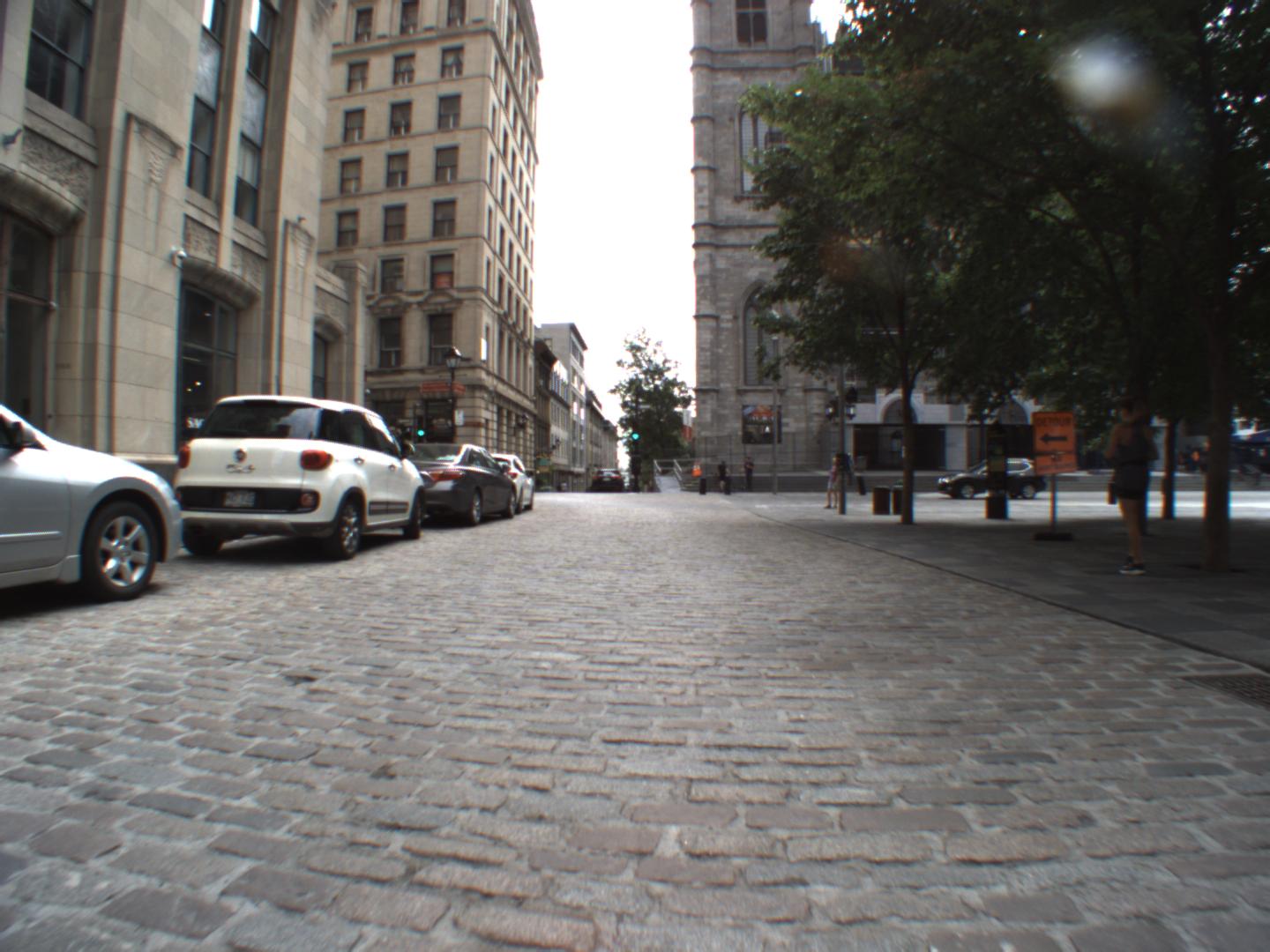}
\includegraphics[width=2.0in]{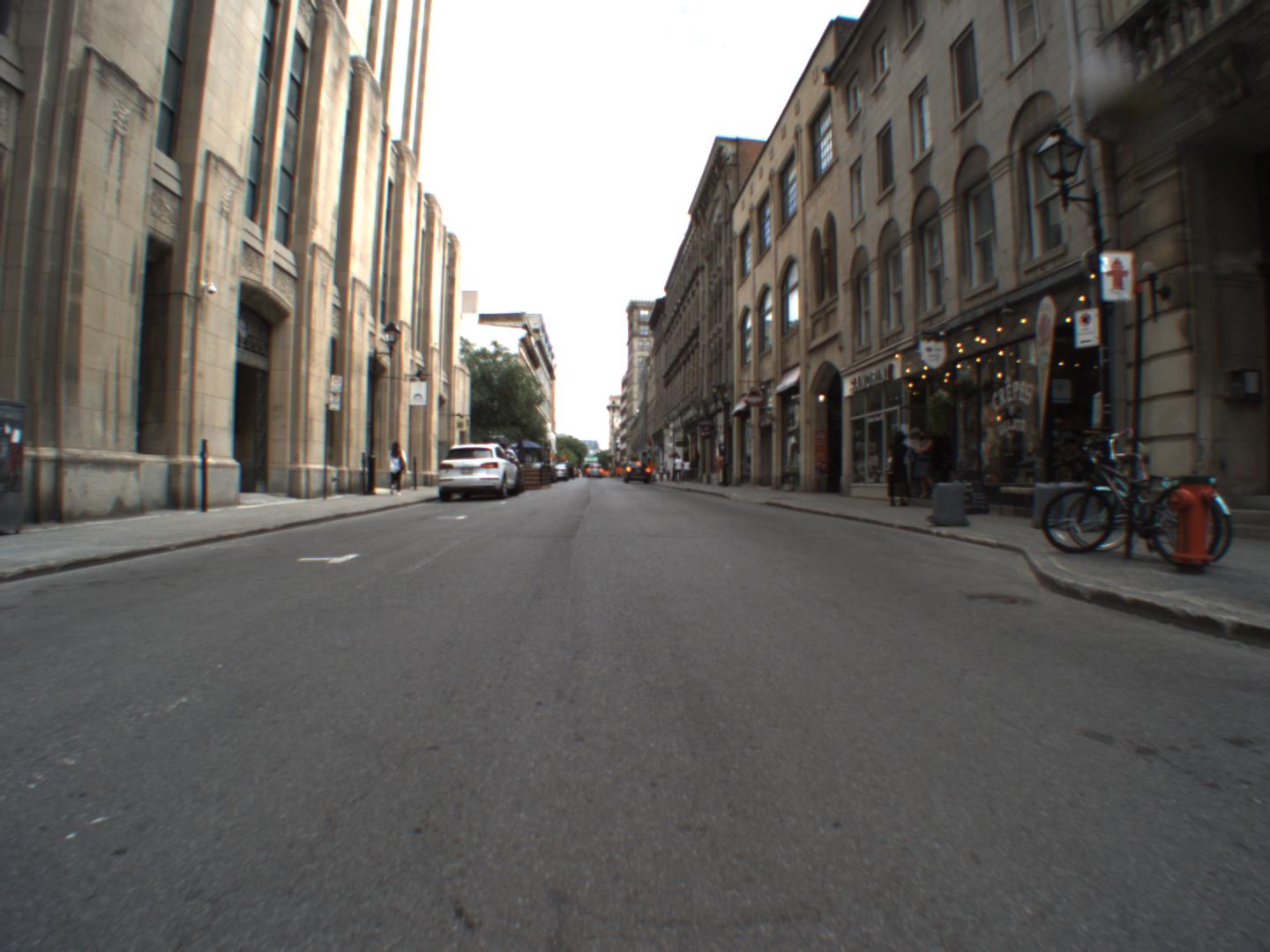}
\caption{Synced forward facing images from LeddarTech \textit{PixSet} (\cite{pixset}), drive 20200721\_144638\_part36\_1956\_2229 through Old Montreal, Quebec.} 
\label{fig:leddartechimg}
\end{figure}

It is important to first demonstrate that the ICET algorithm is capable of robust point cloud registration on realistic data 
before analyzing ICET's estimates of solution error covariance. 
In this section, we validate the performance of our algorithm on a trajectory from the \textit{LeddarTech Pixset} dataset \citep{pixset}, which was chosen for its high-fidelity Ouster OS1-64 LIDAR unit as well as the inclusion of a GNSS/INS fused baseline. The selected trajectory contains 276 frames of movement through an urban environment pictured in Figure \ref{fig:leddartechimg}.

\begin{figure}[h]
\centering
\includegraphics[width=4.5in]{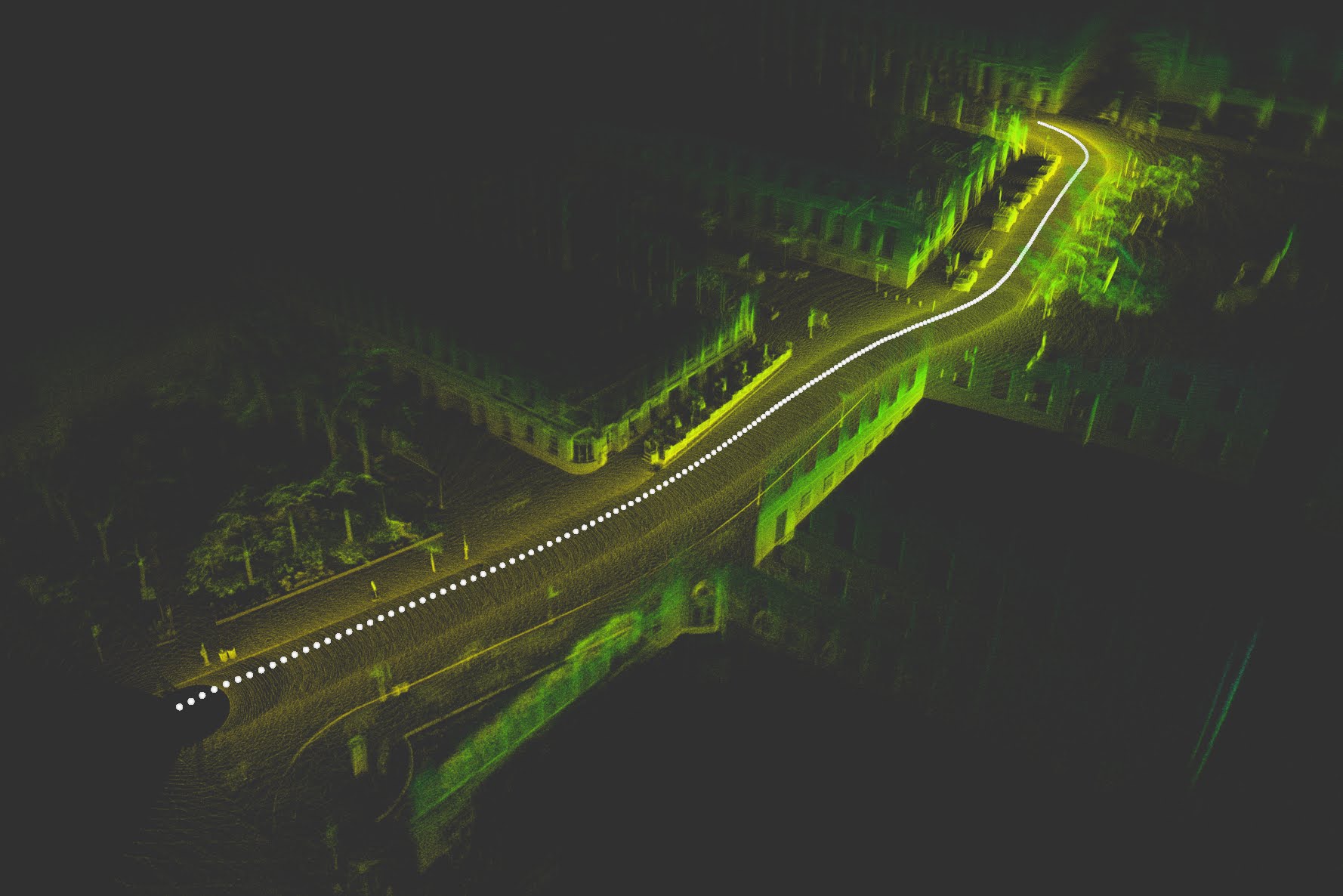}
\includegraphics[width=2.125in]{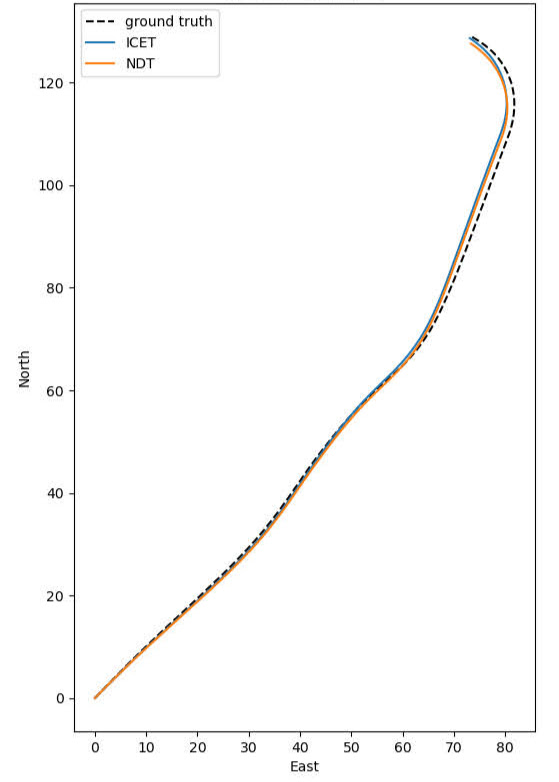}
\caption{ICET demonstration for real-world PixSet \textit{data}. (Left) ICET-odometry trajectory, shown in white, with a High Definition (HD) Map generated using the odometry to stitch scans together. (Right) Vehicle trajectory projected on North-East plane, as measured by three methods: GPS ground truth, ICET odometry, and NDT odometry.}
\label{fig:hdMap}
\end{figure}

For this test, vehicle motion was obtained by estimating the transformations between every sequential pair of frames. No loop closure or graph optimization was implemented for these tests. The resulting trajectory is visualized in Figure \ref{fig:hdMap}.
On the left side of the figure, the trajectory generated by ICET odometry is plotted as a white-dotted curve superposed on an HD map.
This map was constructed using ICET odometry estimates, transforming each successive LIDAR scan by the transformation output at a given step to form a single large point cloud.
The sharp resolution of the map is a strong qualitative indicator of the quality of the ICET odometry. The right side of the figure shows the ICET trajectory on North-East axes, along with GPS\slash INS data and with the trajectory generated by NDT odometry. There is a small but visible difference between the odometry estimates of NDT and ICET. The gap is slightly larger between the LIDAR odometry methods and the GPS data, especially toward the end of the trajectory (upper-right corner). Such divergence over time is expected when comparing dead-reckoning solutions (LIDAR odometry) to absolute positioning (GPS).

In this paper, our primary goal is to assess the quality of the ICET covariance estimate between two sequential frames, as generated by (\ref{eq:Pmatrix}) and conditioned by the sigma-point test. Unfortunately, real-world data are not an effective tool for assessing the quality of our LIDAR covariance predictions. The problem is that the GPS\slash INS data are not sufficient as ground truth when comparing sequential LIDAR frames. The GPS-PPP system produces cm-level localization errors when differencing any two points in time. This becomes a significant advantage when comparing GPS to dead-reckoning over an extended duration (e.g., over the whole 276 frames trajectory shown in Figure \ref{fig:hdMap}); however, for sequential frames, the LIDAR odometry exhibits only mm-level localization errors, an order of magnitude smaller than those for GPS. Because the error in the GPS baseline is too high to provide an effective ground truth when estimating scan-matching error covariance for sequential frames, the next section explores an alternative methodology. In the next section, we 
make use of simulated test environments in which we have access to perfect ground truth values.

\newpage
\section{Verification Testing: High Fidelity Simulated Dataset}\label{HighFidelitySim}

For this test, our goal is to assess scan-matching error for each sequential transformation. We opt to use simulated LIDAR point clouds, which provide direct truth data to which ICET-derived position and attitude changes can be compared. Importantly, 
as other authors have demonstrated \citep{squeezeseg, D2DNDT}, performance of scan registration algorithms on sufficiently realistic synthetic LIDAR point clouds is consistent with performance on similar real-world data, particularly in the case of distributions-based algorithms (like NDT and ICET), as these methods are especially robust to sensor dropout, partial reflection, and other artifacts observed in field applications.

To this end, 
our second simulation environment is a high-fidelity urban scene. Specifically we use data from the \textit{Cooperative Driving Datatset} (CODD) \citep{CODD}, which was generated in the \textit{CARLA} open-source driving simulator \citep{CARLA}. The testing environment is representative of a city with detailed features absent in the abstracted simulations conducted in the following section. Included are loose foliage, complex organic-shaped structures, and moving objects like pedestrians and vehicles.
This high-fidelity data set considers a series of LIDAR locations along a simulated automotive trajectory through an urban landscape, and introduces randomness due to sensor noise as well as different physical geometries for each scan. In the CODD data, measurement noise is dominated by range (time of flight) error, and is set to modest but representative levels. Specifically range errors are modeled as normally distributed with a standard deviation of 1cm. 

\begin{figure}
\centering
\includegraphics[width=4.0in]{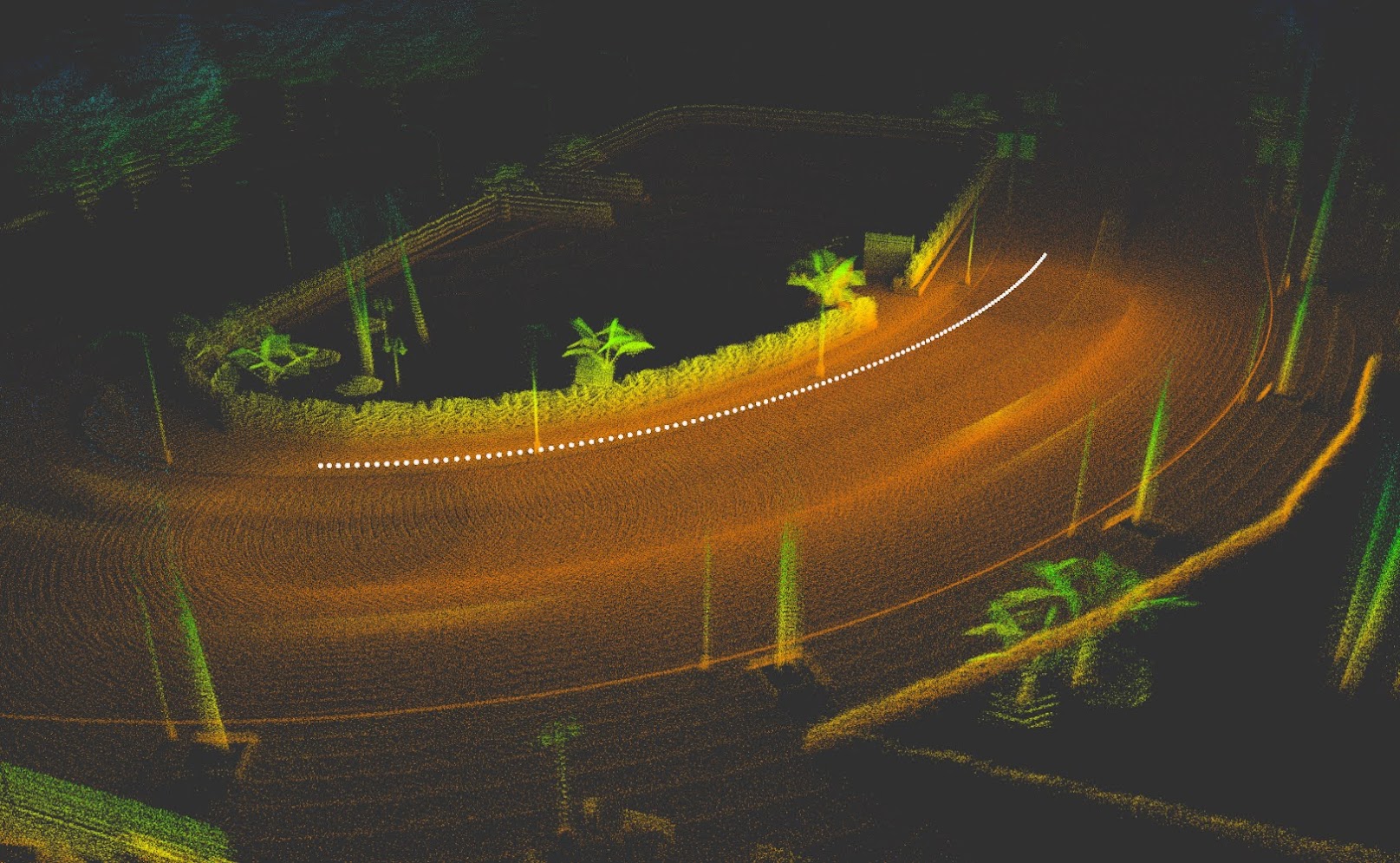}
\includegraphics[width=2.0in]{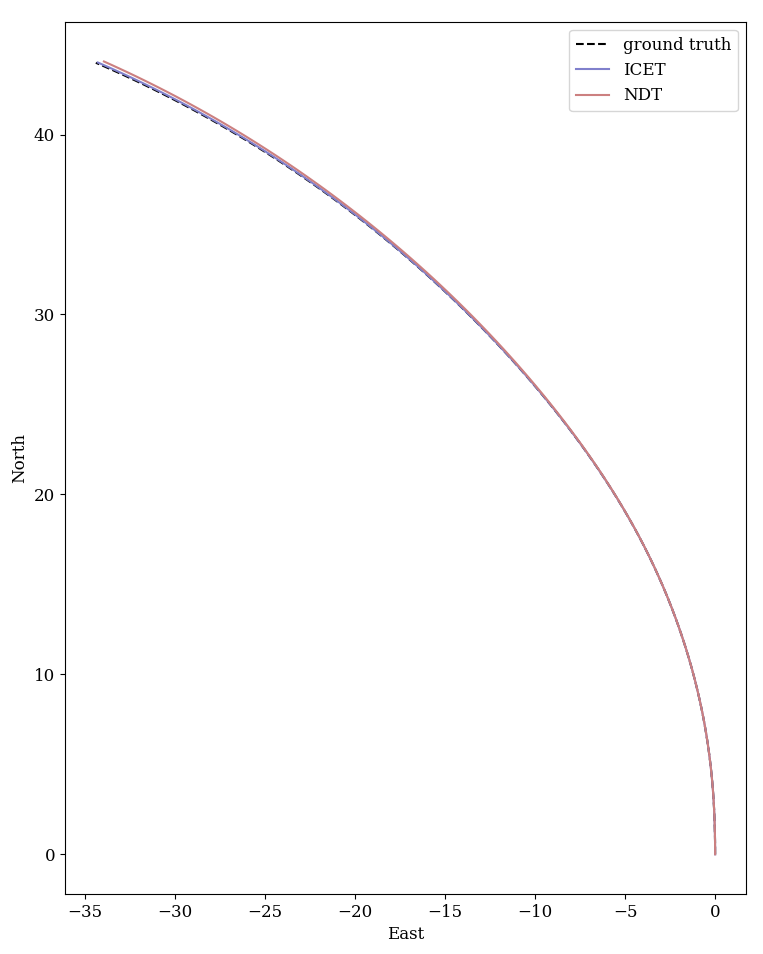}
\caption{ICET performance for simulated CODD data. (Left) ICET-odometry trajectory, shown in white, with an HD Map generated using the odometry to stitch scans together. (Right) Vehicle trajectory projected on North-East plane, including the known truth from the simulation (dashed line) as well as ICET and NDT odometry measurements (solid lines).}
\label{fig:codd}
\end{figure}

The initial condition for each new scan is defined by zeroing the transformation vector (\ref{eq:xvec}), which is equivalent to starting with the assumption that there is no motion between subsequent scans.
We parsed the data in the CODD dataset into 123 sequential (overlapping) scan pairs, which represented the full driving trajectory for one CODD vehicle recording data at 5 Hz over a period of 24.6 seconds. Figure \ref{fig:codd} visualizes the CODD data, showing an HD map generated from the ICET odometry data as well as an overhead plot of the estimated trajectories for NDT and ICET compared against truth.

\begin{figure}
\centering
\includegraphics[width=3.14in]{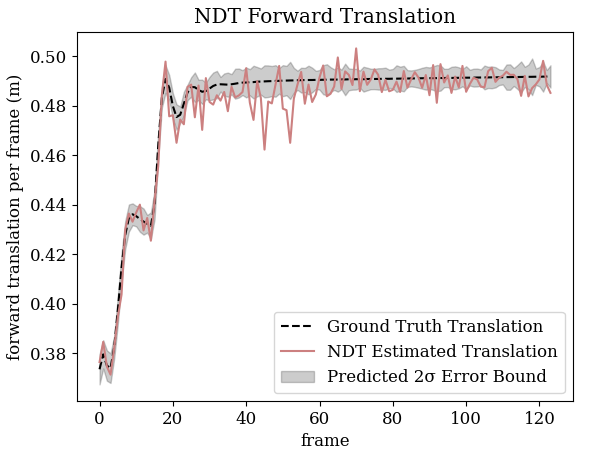}
\includegraphics[width=3.14in]{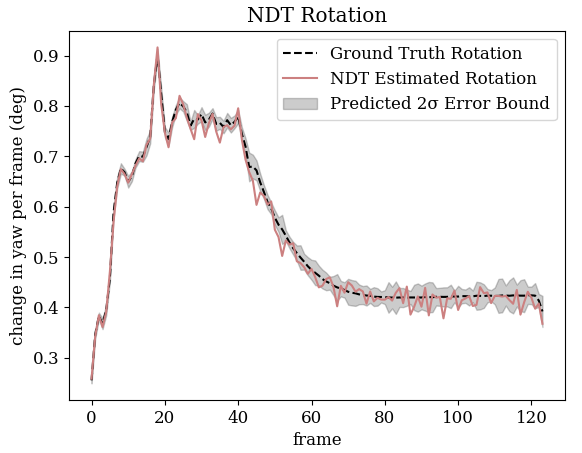}
\includegraphics[width=3.14in]{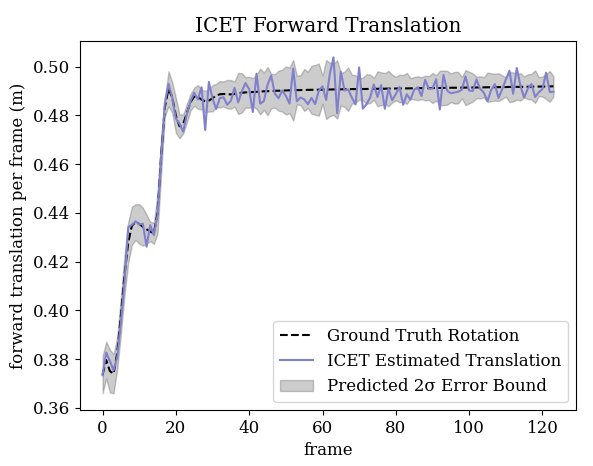}
\includegraphics[width=3.14in]{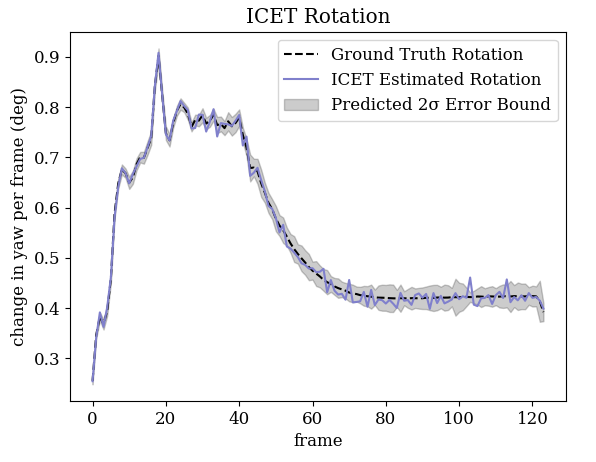}
\caption{NDT and ICET generated estimates for forward translation and rotation about the vertical axis compared to ground truth values. Shaded regions show the 2$\sigma$ error bound for each frame predicted by each algorithm.}
\label{fig:accum}
\end{figure}

For this test, we applied our full ICET algorithm (including the sigma-point refinement) as well as an NDT baseline (without sigma-point refinement) to each sequential pair of LIDAR scans. To provide a fair comparison, NDT and ICET used the same set of hyperparameters including voxel size 
(4${}^\circ$ resolution in azimuth and elevation), voxel boundaries, number of iterations, moving-object rejection threshold, and minimum number of points per cell (50 points). 
For this study, the key parameter is voxel size. As discussed in Section \ref{Problem}, voxels must be big enough to capture random errors, yet not so big as to envelope large objects. For this reason, the Appendix performs a voxel-size parameter study. The Appendix shows little sensitivity in performance over a wide range of voxel sizes (up to a breaking point when performance quickly degrades if voxels are too large or too small).

Since we have known truth from the simulation, we can compute the true motion between each sequential scan and compare to the odometry solution. These frame-to-frame motions are plotted in Figure (\ref{fig:accum}). The figure contains a grid of four plots, with the first column showing frame-to-frame changes in along-track position and with the second column showing frame-to-frame changes in yaw heading. The top row shows NDT results, and the bottom row shows ICET results. In the figures, the dahsed black line indicates simulated truth, the solid line indicates the odometry, and the shaded region indicates the 2$\sigma$ error bound about the truth. The 2$\sigma$ error bound, is computed from (\ref{eq:Pmatrix}), and differs slightly between NDT and ICET because the voxel-level covariance estimates are different in each case.

Notably, the ICET bounds are more representative of the actual errors than the NDT bounds.
For both algorithms, Figure \ref{fig:accum} indicates that the predicted 2$\sigma$ error bounds grow and shrink as the vehicle passes through the simulated environment, however, the change in bound size throughout the trajectory is far more pronounced in ICET than in NDT. In ICET, the widest translational error bounds fall between frames 40 and 70, where 
there are large extended surfaces (fences and foliage) parallel to the road on both sides of the vehicle.
The ICET bounds then come back down in size after frame 70, presumably because the vehicle rounds the corner where small-scale features become visible to the LIDAR sensor. In contrast, NDT's estimated error bounds are relatively consistent in magnitude throughout the same range, presumably because it is mis-characterizing ambiguous LIDAR measurements along large surfaces and integrating them into the solution and covariance estimate.
In theory, the 2$\sigma$ bounds should contain roughly 95\% of the estimated states.
For NDT, the bounds only contain the translation error 40\% of the time (73 out of 123 instances); by comparison, the ICET bounds contain the translation error 93\% of the time (115 of 123 instances). The results are similar for rotation error, with NDT bounds containing only 77\% of the time (95 of 123 instances) and with ICET bounding 95\% of the time (117 of 123 instances). In short, ICET's 
 predicted covariance metrics are reasonably representative of true solution error on realistic point cloud data, while those provided by NDT are not. 
\section{Verification Testing: Abstracted Geometries}\label{AbstractSim}

As a final test, we consider the corner-case where ICET's sigma-point refinement might remove \textit{all} of the data in one or more solution coordinates. This might happen, for example, in a straight and featureless tunnel, where there is no valid LIDAR measurement data in the along-track direction. In such a case, we expect ICET should still function, and moreover that it should enunciate that the solution is unavailable in the ambiguous direction. It is notable the issue of insufficient LIDAR scan-matching data along one or more axes has also recently been identified as a problem for other algorithms, too \citep{xicp}). 


\begin{figure}[h]
\centering
\includegraphics[width=2.0in]{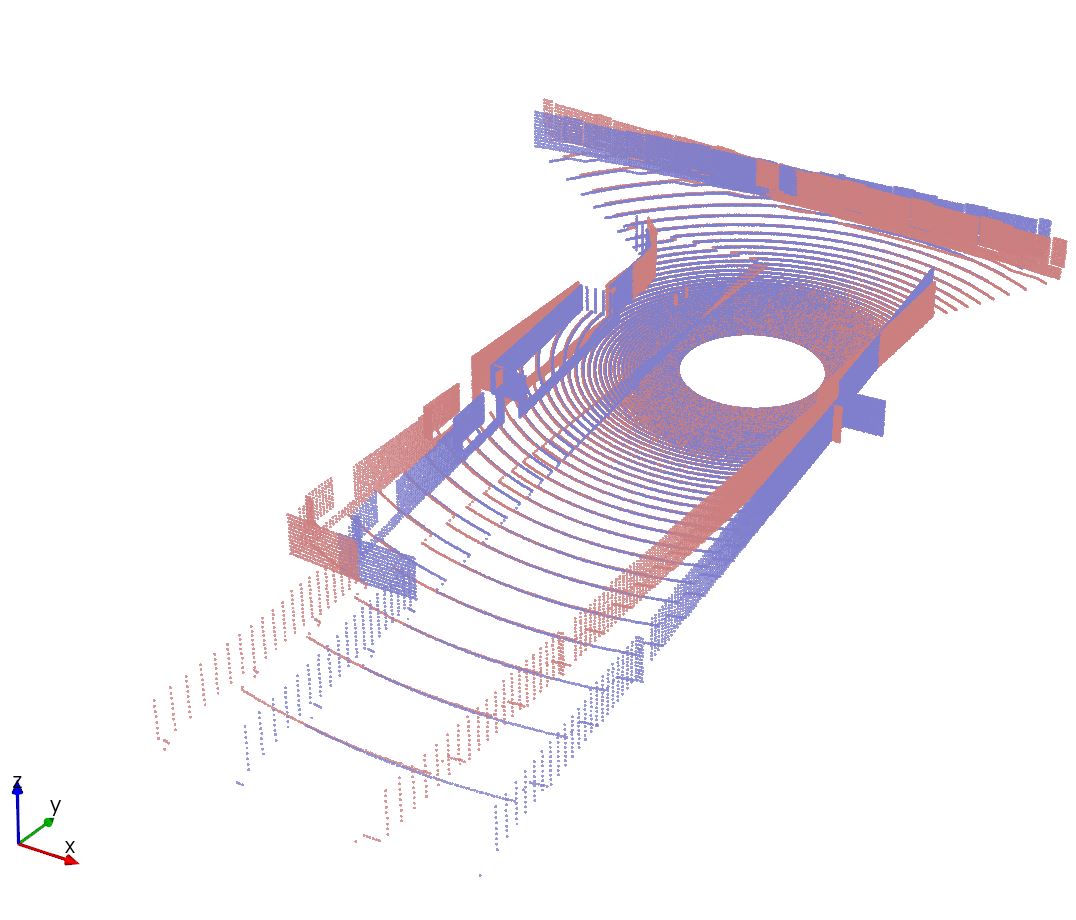}
\includegraphics[width=1.5in]{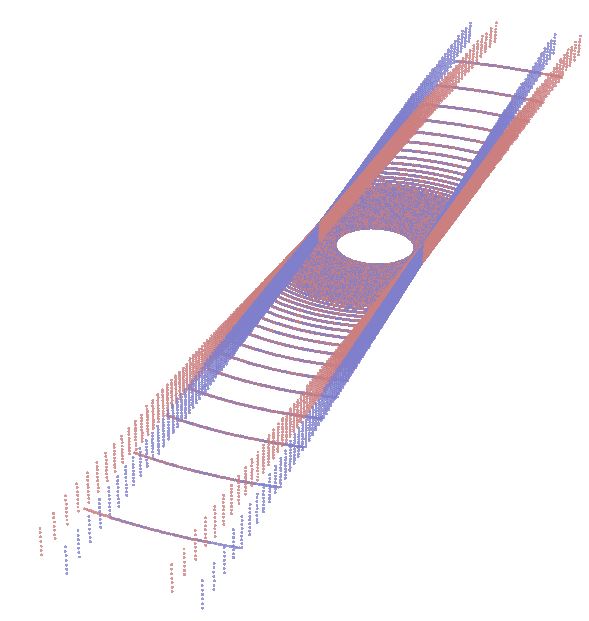} 
\includegraphics[width=2.0in]{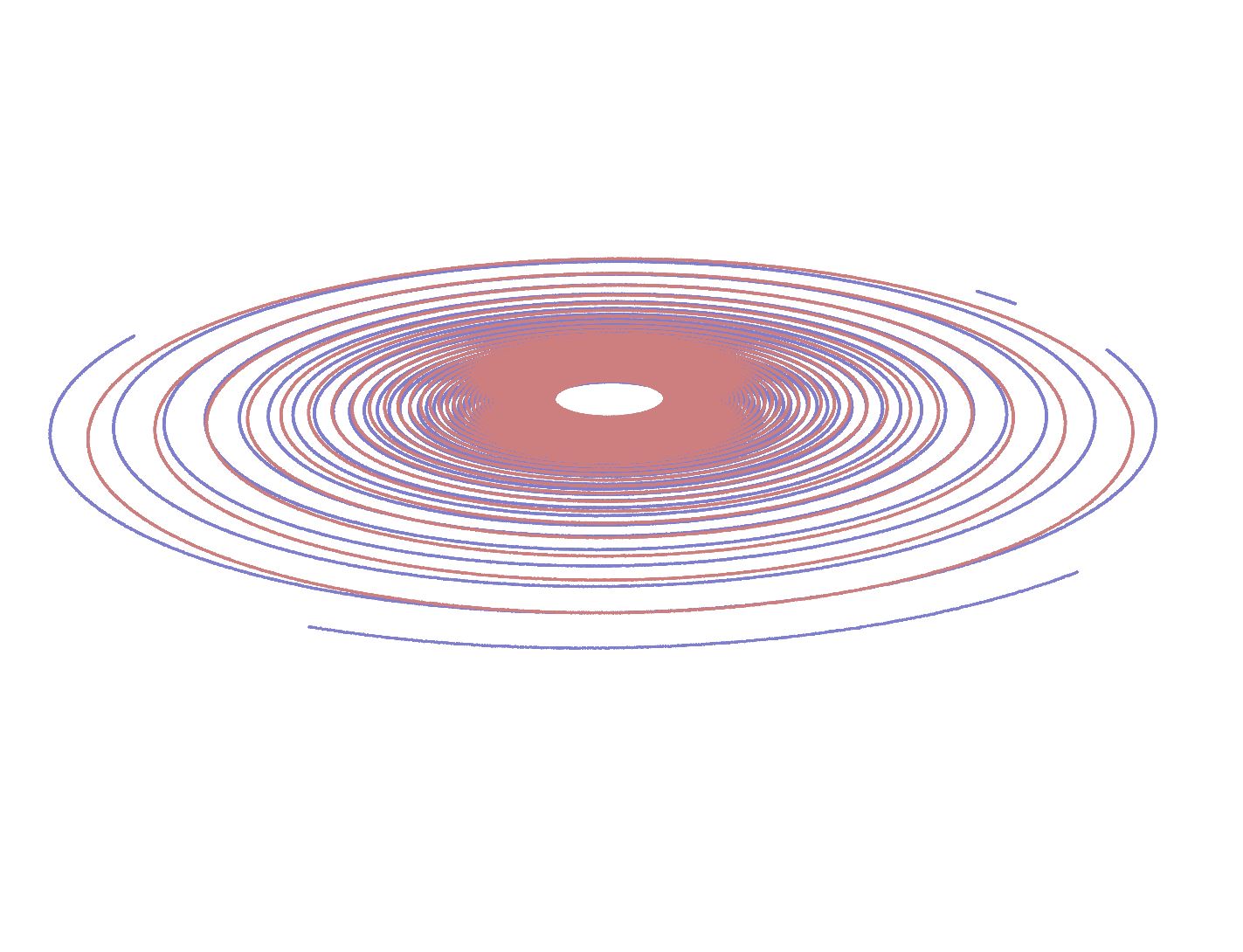}
\caption{Corner-case simulations including scans for (left) a T-intersection, (middle) a tunnel, and (right) a large open field.}
\label{fig:threeRoads}
\end{figure}

We consider three corner-cases, each constructed from an abstracted geometry. The geometries include a T-intersection, a straight tunnel, and an open field. In each case, we synthesize LIDAR measurements via the Matlab Automated Driving Toolbox. The three geometries are depicted in Figure \ref{fig:threeRoads}. For each geometry, the figure shows a pair of scans: a reference scan (blue) and a perturbed scan (red) that must be aligned to the reference scan. In using Matlab to generate these scans, we applied zero-mean Gaussian noise with an 0.2 cm standard deviation in each of the \textbf{x}, \textbf{y}, and \textbf{z} directions.

The abstracted geometries were designed specifically to check ICET's sigma-point exclusion and rank-testing capabilities, as described in Equation (\ref{eq:cond1}). The T-intersection consists of extended walls in all directions. No large-scale ambiguity exists, since data exist to constrain all six transformation states, including the translation estimates in the $x$, $y,$ and $z$ directions and rotation estimates in $\phi$, $\theta$ and $\psi$. By contrast, the straight tunnel provides useful information in all translation and rotation directions except in the road-aligned (or $y$) direction. The flat plane provides useful information only in the vertical, pitch, and roll directions, and not in the horizontal and yaw directions. We hypothesize that the the sigma-point test should detect walls (and remove wall-coincident measurements) in all three simple scenes; however, only the latter two scenes are expected to fail the rank-checking test related to (\ref{eq:cond1}). Note that we avoid edge effects by sampling the reference and new scans away from the boundaries of the simulated domain.  

In these abstracted geometry tests, we do not create a full trajectory; instead we repeatedly generate new noise for the two scans via a Monte Carlo simulation. 
For each Monte Carlo trial, the initial guess is generated from a zero-mean, independent Gaussian distributions with a standard devation of 12.5cm in $\{x,y,z\}$ and with a standard devation of 1.7${}^\circ$ in $\{\phi,\theta,\psi\}$. In all, we generated 500 Monte Carlo trials for each abstracted geometry, each with 
a unique seed for measurement noise and randomized initial translation and rotation.

For each Monte Carlo trial, we ran two algorithms to conduct scan matching: the full ICET algorithm as well as 
a comparable implementation of NDT, similar in all respects to the ICET algorithm save for the sigma-point refinement and the related condition-number test of (\ref{eq:cond1}).
Again, the voxel grid was defined with $4^\circ$ resolution in azimuth and elevation, with a minimum number of points per voxel of 50.
Error statistics were compiled over all trials, with error computed as the difference of each state compared to the truth.  Root-mean-square error (RMSE) values are plotted in Table (\ref{tab:t-int}) for the T-intersection, Table (\ref{tab:tunnel}) for the tunnel, and Table (\ref{tab:plane}) for the plane.  The tables list the results for the NDT baseline (first two rows of data) followed by the results for the ICET implementation (last two rows of data). Statistical RMSE values are reported in the table, alongside predicted values,  extracted from the diagonal of $\mathbf{P}$ as computed analytically by (\ref{eq:Pmatrix}).  

\vspace*{-10pt}

\begin{center}
\begin{table*}[h]%
\centering
\caption{T-Intersection Case 
}%
\rowcolors{2}{gray!25}{white}
\begin{tabularx}{\textwidth}{@{\hspace*{6pt}}l X X X X X X@{\hspace*{6pt}}}
\toprule
\rowcolor{gray!50}
& $x$ (cm) & $y$ (cm) & $z$ (cm) & $\phi$ (deg) & $\theta$ (deg) & $\psi$ (deg) \\
\midrule
NDT: RMSE & 0.0128 & 0.101 &  0.0158 & 0.00131 & 0.00218 & 0.00104 \\ 
NDT: Predicted  & 0.0122 & 0.0522 & 0.0147 & 0.00127 & 0.00199 & 0.00102\\
\midrule
ICET: RMSE & 0.0122 & 0.0709 &  0.0159 & 0.00134 & 0.00213 & 0.00103 \\ 
ICET: Predicted & 0.0124 & 0.0736 & 0.0152 & 0.00129 & 0.00215 & 0.00105\\
\bottomrule
\end{tabularx}
\begin{tablenotes}
\end{tablenotes}
\label{tab:t-int}
\end{table*}
\end{center}

\vspace*{-40pt}

\begin{center}
\begin{table*}[h]%
\centering
\caption{Straight Tunnel Case 
}%
\rowcolors{2}{gray!25}{white}
\begin{tabularx}{\textwidth}{@{\hspace*{6pt}}l X X X X X X@{\hspace*{6pt}}}
\toprule
\rowcolor{gray!50}
& $x$ (cm) & $y$ (cm) & $z$ (cm) & $\phi$ (deg) & $\theta$ (deg) & $\psi$ (deg) \\
\midrule
NDT: RMSE & 0.0182 & 10.493* &  0.0113 & 0.00147 & 0.00148 & 0.000987 \\ 
NDT: Predicted & 0.0164 & 0.0655 & 0.0106 & 0.00140 & 0.00142 & 0.000881\\ 
\midrule
ICET: RMSE & 0.0171 & Marked DNU &  0.0110 & 0.00144 & 0.00151 & 0.000920 \\ 
ICET: Predicted & 0.0170 & Marked DNU & 0.0107 & 0.00143 & 0.00153 & 0.000901\\
\bottomrule
\end{tabularx}
\begin{tablenotes}
*Output error unbounded for this solution component
\end{tablenotes}
\label{tab:tunnel}
\end{table*}
\end{center}


\begin{center}
\begin{table*}[h]%
\centering
\caption{Open Field Case 
}%
\rowcolors{2}{gray!25}{white}
\begin{tabularx}{\textwidth}{@{\hspace*{6pt}}l X X X X X X@{\hspace*{6pt}}}
\toprule
\rowcolor{gray!50}
& $x$ (cm) & $y$ (cm) & $z$ (cm) & $\phi$ (deg) & $\theta$ (deg) & $\psi$ (deg) \\
\midrule
NDT: RMSE & 10.447* & 10.424* &  0.0116 & 0.00127 & 0.00128 & 1.73* \\
NDT: Predicted & 0.181 & 0.181 & 0.0115 & 0.00127 & 0.00128 & 0.0121\\
\midrule
ICET: RMSE & Marked DNU & Marked DNU &  0.0117 & 0.00128 & 0.00127 & Marked DNU \\
ICET: Predicted & Marked DNU & Marked DNU  & 0.0115 & 0.00128 & 0.00128 & Marked DNU \\
\bottomrule
\end{tabularx}
\begin{tablenotes}
*Output error unbounded for this solution component
\end{tablenotes}
\label{tab:plane}
\end{table*}
\end{center}


As expected, ICET's sigma-point refinement excludes measurements and when all measurements in a given coordinate direction are lost, the condition number test excludes those axes from the solution. This result can be seen clearly from the tables.
In cases where the condition-number test removes dimensions from state estimation, those dimensions are labeled in the table as ``\textit{Marked DNU}'' where \textit{DNU} stands for ``Do Not Use.'' 
The table indicates that ICET correctly excluded the $y$ component in the tunnel (for all Monte Carlo trials), while NDT confidently reported incorrect estimates. Similarly, in the open-field scene, ICET successfully excluded all $x$, $y$, and $\psi$ solution components.

The standard deviations for scan-matching error that are reported in the table are quite favorable (sub-millimeter translation errors and millidegree rotation errors). It is relevant to note that these error levels are consistent with the results reported in prior NDT simulation studies, notably in \citep{D2DNDT}. Low scan-match errors are critical for LIDAR odometry applications, where the standalone dead-reckoning error is an accumulation of the scan-match errors over time.

Importantly, the tables show clearly that ICET can reliably predict accuracy in these scenes and that the prediction is improved subtly by the sigma-point refinement step. With no refinement,  $\mathbf{P}$ consistently underpredicts the error.  Consider the T-intersection data in Table (\ref{tab:t-int}), where for each state, the predicted standard deviation is lower than the actual standard deviation for the NDT case, with the percent difference between -2\% (for $\psi$) and -10\% (for $\theta$) except in one extreme case with relative error of -93\% (for $y$).  By comparison, the predictions for ICET are neither consistently above nor below the actual values, under-predicting $\phi$ and $z$ by -3.9\% and -4.6\%, and slightly over-predicting for the other cases, up to +4.2\% (for $y$).

Although the effect of the sigma-point refinement is generally somewhat subtle for the T-intersection, the difference in the $y$ (along-track) direction is truly striking. Table (\ref{tab:t-int}) shows that the NDT case (with no extended surface suppression) underpredicts the $y$ standard deviation by a factor of nearly 2. The issue here is that the majority of voxels contain walls aligned with the along-track direction.  Even though the measurements in these directions have a somewhat large sample covariance (driven by voxel width), these large-covariance voxels are not significantly deweighted by $\mathbf{W}$, because there are very few voxels with "good" information in the along-track direction. As a result, NDT becomes overconfident in its along-track estimate.  By comparison, the sigma-point refinement entirely removes along-wall measurements, so they are not even considered in the weighting matrix or the computation, resulting both in lower error and a better prediction of that error.

The effect of sigma-point refinement is even more pronounced in the straight-tunnel and open-field scenes, where strong ambiguities were present.  In these scenes, the full ICET algorithm reports that the solution quality is poor where the ambiguity exists (as denoted by the "Excluded" cells in the tables). Without the refinement, NDT produces a somewhat arbitrary error that is highly correlated to the initial condition.  Recall that the initial-condition error was 12.5 cm in translation and 1.7 degrees in rotation, which is reflected in the starred values in the tables (10.4 cm in translation and 1.7 degrees in translation). In these cases, the sigma-point refinement plays a critical role by recognizing and reporting ambiguity.  As a side-effect, the sigma-point refinement also converges more quickly (for the open field case: within 5 iterations for full ICET as opposed to 12 iterations for NDT).

Sigma-point refinement both reduces estimation error and improves the ability of the algorithm to predict the accuracy of its solution. The degree to which the true and predicted accuracy of the algorithm is improved is a function of the geometry of the scene and the size of the voxels. For instance, in the T-Intersection case, there are enough distributions artificially limited in the $y$ direction by voxel width that the sigma-point refinement greatly impacts both the actual and predicted error in the $y$ direction. In the same scene, however, the accuracy and accuracy prediction in the $z$ direction are very similar whether or not the sigma-point refinement is employed. 

\section{Conclusion}\label{conclusion}

\if false

In the Figure \ref{fig:pancakes}, the same planar surface is characterized twice, using different size voxel grids to form distributions of points. For coarse voxels, it is obvious that the distributions are much more oblate (pancake-shaped) while the finer grid cells produce much rounder distributions. This example highlights the downsides associated with smaller voxel sizes when not mitigating for locally ambiguous surfaces.
Fine grids allow the Gaussian Mixture Model \cite{GMM} which underpins all distribution-based registration techniques better avoid artifacts of grid discretization by providing distinct regions of a frame to each unique face of an object.
Voxels must be sufficiently small such that they are capable of properly characterizing a region with a single ellipsoid, however, as is visible in this figure, as voxel sizes shrink, so do the maximum possible lengths of distribution ellipsoids aligned in the direction of an ambiguous surface.  
The attraction of distribution centers in directions subject to the aperture ambiguity problem is somewhat mitigated in NDT by the way in which the \textit{Mahalanobis Distance} is used to weigh D2D residuals.

\begin{figure}[h]
\centering
\includegraphics[width=2.0in]{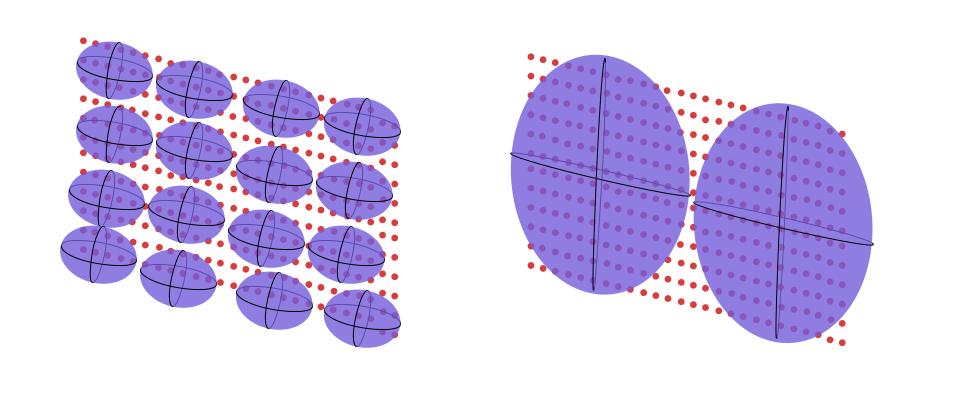}
\caption{Effect of changing voxel size on proportion of ellipsoids used to characterize a common surface}
\label{fig:pancakes}
\end{figure}

For large voxel sizes, planar surfaces produce highly oblate covariance ellipsoids which are extremely compact in the useful direction normal to the surface, and very extended in the other directions. This causes any residuals in the compact (useful) directions of each voxel to be weighted much more heavily when calculating that voxel's contribution to the solution vector than residuals in any other direction. This trick starts to fall apart as the size of voxels decreases and the covariance ellipsoids become less oblate. At smaller voxel sizes, the algorithm is tricked into thinking that extended directions of per-voxel distributions are more useful than the same useless directions on the same surfaces would be if using larger voxels-- this is because the size of voxels puts an artificial bound on the degree to which ambiguous surfaces can be de-weighted. An unfortunate property of other distributions-based algorithms, such as vanilla NDT, is their tendency to underestimate the solution vector for point clouds in the final stages of convergence. At smaller sizes, the registration process behaves similarly to an overdamped, consistently undershooting the true solution. Smaller voxel sizes are often required to properly characterize surfaces containing complex curves, as a single ellipsoid can not adequately describe the features. 
\fi


A topic for future work is to identify and address large-scale ambiguities that may appear in LIDAR scenes. While our proposed strategy is capable of identifying local structure and global ambiguity that results in a rank deficiency, other types of ambiguity are not directly addressed by ICET. Most notably, ICET does not flag scenes containing a lattice of repeated features (e.g. driving past regularly spaced columns).  Such a scene would result in an \textit{integer ambiguity} problem, much like the problem of resolving wavelength in carrier-phase processing for GNSS \citep{enge1994global}. Ambiguity may also be masked by the WLS linearization. For instance, if a LIDAR is mounted on a vehicle moving along a curved tunnel, the along-track translation and yaw rotation should be coupled and an ambiguous combination of states should be identified by the condition-number test; however, the condition-number test may not recognize this ambiguity if the curve is sufficiently sharp. New algorithm extensions will be needed to identify such challenging cases of global ambiguity.

The main result of this paper was the introduction of \textit{Iterative Closest Ellipsoidal Transformation} (ICET), an algorithm for scan-matching with 3D LIDAR point-cloud data. As a LIDAR processing algorithm, ICET is similar to the well known \textit{Normal Distributions Transform} (NDT) algorithm, with the addition of a capability to recognize differences between randomness and deterministic structure in voxel point distributions and to leverage that information to construct a more representative model of the solution-error covariance. 
ICET is formulated 
using a least-squares method, supplemented by a sigma-point test to identify point distributions that are dominated by deterministic structure rather than measurement randomness. The 3D ICET algorithm was applied to real data (to demonstrate proof-of-concept), high-fidelity simulations (to study error properties with perfect ground truth), and corner-case simulation (to demonstrate the significance of dimension reduction for the solution using a condition-number test).


\section*{Acknowledgments}
The authors wish to acknowledge and thank the U.S. Department of Transportation Joint Program Office (ITS JPO) and the Office of the Assistant Secretary for Research and Technology (OST-R) for sponsorship of this work. We also gratefully acknowledge NSF grant CNS-1836942, which supported specific aspects of this research. Opinions discussed here are those of the authors and do not necessarily represent those of the DOT, NSF, or other affiliated agencies.



\section*{Conflict of interest}

The authors declare no potential conflict of interests.

\bibliography{references}%

\appendix

\section{Parameter Study: Spherical Voxel Resolution}\label{parameterstudy}

As noted in Section \ref{AbstractSim}, voxel size is a key parameter, since the sigma-point algorithm will not function correctly if voxels are too small to capture the random error distribution or too large to exclude extended objects effectively. Given the importance of voxel size to our concept, this Appendix performs a sensitivity study over a range of voxel sizes. 

\begin{figure}[h]
\centering
\includegraphics[width=3.2in]{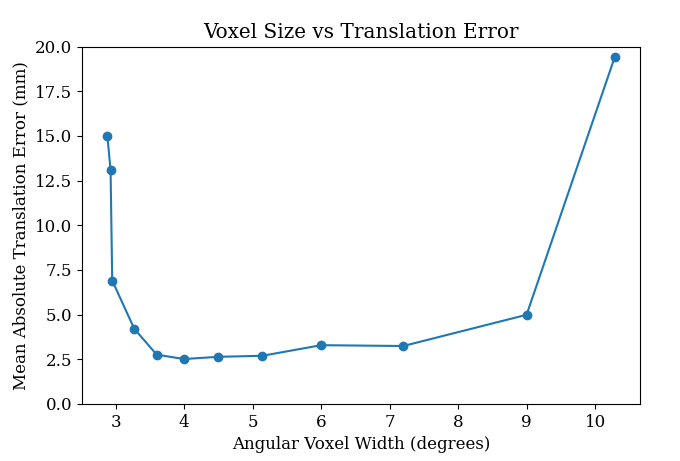}
\caption{Sensitivity study: Performance as a function of voxel resolutions}
\label{fig:voxelsize}
\end{figure}

In this parameter-sensitivity study, we consider use the CODD dataset to characterize error performance as a function of voxel size. The voxel grid is a one-cell deep spherical grid, where the radial boundaries of each cell are adapted to the point distribution. The number of voxels is governed by the angle subtended by each voxel.  In this sensitivity study, we consider angles between 2.7$^\circ$ and 10$^\circ$. Solution error values were computed over the full trajectory studied in Section \ref{AbstractSim}, and mean-absolute error values were tabulated as a function of resolution. Figure \ref{fig:voxelsize} plots error values (mm) as a function of resolution (degrees of voxel width). The figure
 demonstrates that the effect of changing the resolution is relatively small over a wide range of angles. Between $3.5^\circ$ and $7^\circ$ the curve is nearly flat, with mild curvature resulting in a minimum error for a resolution of $4^\circ$.

Although performance is a very weak function of voxel width over a wide range of resolutions, the figure shows a dramatic loss of performance at the extremes, when voxel width drops below $3^\circ$ or climbs above $9^\circ$.
 When voxel resolution grows larger than $9^\circ$, presumably the cells are so coarse that they fit distributions to multiple surfaces occupying the same voxel, thereby treating deterministic scatter in data points as random noise.
 When voxel resolution is set to less than $3.5^\circ$, voxels struggle to capture the extend of randomness and, perhaps more importantly, voxels become data starved because they cannot capture points from more than one scan line at a time. As such, we would expect less sensitivity on the left side of the plot if we were to use a LIDAR sensor with more beams to provide higher resolution across elevation angles. 

\end{document}